\documentclass{article}

\usepackage{microtype}
\usepackage{graphicx}
\usepackage{subfigure}
\usepackage{booktabs} % for professional tables
\usepackage{mathtools}

\newcommand{\R}{ \mathbb{R}}
\usepackage[colorinlistoftodos,bordercolor=orange,backgroundcolor=orange!20,linecolor=orange,textsize=scriptsize]{todonotes}

\newcommand{\squeeze}{\textstyle}
\usepackage{colortbl}
\definecolor{bgcolor}{rgb}{1, 0.86, 0.69}
\usepackage{hyperref}

\usepackage[accepted]{icml2021}

\icmltitlerunning{Random Reshuffling with Variance Reduction: New Analysis and Better Rates}
\usepackage{amsmath}
\usepackage{amsfonts}
\usepackage{amsthm}
\usepackage{mathtools}

\newcommand{\eqdef}{\coloneqq}
\newtheorem{proposition}{Proposition}
\newtheorem{definition}{Definition}
\newtheorem{theorem}{Theorem}
\newtheorem{corollary}{Corollary}
\newtheorem{lemma}{Lemma}
\newtheorem{assumption}{Assumption}

\begin{document}

\twocolumn[
\icmltitle{Random Reshuffling with Variance Reduction: New Analysis and Better Rates}




\begin{icmlauthorlist}
\icmlauthor{Grigory Malinovsky}{KAUST}
\icmlauthor{Alibek Sailanbayev}{KAUST}
\icmlauthor{Peter Richt\'arik}{KAUST}

\end{icmlauthorlist}

\icmlaffiliation{KAUST}{King Abdullah University of Science and Technology (KAUST)}

\icmlcorrespondingauthor{Grigory Malinovsky}{https://grigory-malinovsky.github.io}

\icmlkeywords{Machine Learning, ICML}

\vskip 0.3in
]



 % otherwise use the standard text.
 \printAffiliationsAndNotice{} 

\begin{abstract}
Virtually all state-of-the-art methods for training supervised machine learning models are variants of SGD enhanced with a number of additional tricks, such as minibatching, momentum, and adaptive stepsizes. One of the tricks that works so well in practice that it is used as default in virtually all widely used machine learning software is {\em random reshuffling (RR)}. However, the practical benefits of RR have until very recently been eluding attempts at being satisfactorily explained using theory. Motivated by recent development due to Mishchenko, Khaled and Richt\'{a}rik (2020), in this work we provide the first analysis of SVRG under Random Reshuffling (RR-SVRG) for general finite-sum problems. First, we show that RR-SVRG converges linearly with the rate $\mathcal{O}(\kappa^{3/2})$ in the strongly-convex case, and can be improved further to $\mathcal{O}(\kappa)$ in the big data regime (when $n > \mathcal{O}(\kappa)$), where $\kappa$ is the condition number. This improves upon the previous best rate $\mathcal{O}(\kappa^2)$ known for a variance reduced RR method in the strongly-convex case due to Ying, Yuan and Sayed (2020). Second, we obtain the first sublinear rate for general convex problems. Third, we establish similar fast rates for Cyclic-SVRG and Shuffle-Once-SVRG. Finally, we develop and analyze a more general variance reduction scheme for RR, which allows for less frequent updates of the control variate. We corroborate our theoretical results with suitably chosen experiments on synthetic and real datasets. 
\end{abstract}

\section{Introduction}
The main paradigm for training supervised machine learning models---Empirical Risk Minimization (ERM)---is an optimization problem of the finite sum structure \begin{equation}
\label{eq:main_finite_sum}
\squeeze
\min \limits_{x \in \R^d} f(x) \eqdef  \frac{1}{n}\sum \limits_{i=1}^n f_i(x),
\end{equation}
where $x \in \R^d$ is a vector representing the parameters (model weights, features) of a model we wish to train, $n$ is the number of training data points, and $f_i(x)$ represents the (smooth) loss of the model on data point $i$. The goal of ERM is to train a model whose average loss on the training data is minimized. This abstraction allows to encode virtually all supervised models trained in practice, including linear and logistic regression and neural networks. 

The gigantic size of modern training data sets necessary to train models with good generalization poses severe issues for the designers of methods for solving \eqref{eq:main_finite_sum}. Over the last decade, stochastic first-order methods have emerged as the methods of choice, and for this reason, their importance in machine learning remains exceptionally high~\cite{Bottou2018}. 
Of these, stochastic gradient descent (SGD) is perhaps the best known, but also the most basic. SGD has a long history~\cite{robbins1951stochastic,bertsekas1996neuro} and is therefore well-studied and well-understood~\cite{Rakhlin2012,hardt2016train,Drori2019,gower2019sgd,Nguyen2020}. 

{\bf Variance reduction.} Despite its simplicity and elegance, SGD has a significant disadvantage: the variance of naive stochastic gradient estimators of the true gradient remains high throughout the training process, which causes issues with convergence. When a constant learning rate is used in the smooth and strongly convex regime, SGD converges linearly to a neighborhood of the optimal solution of size proportional to the learning rate and to the variance of the stochastic gradients at the optimum~\cite{gower2020variance}. While a small or a decaying learning schedule restores convergence, the convergence speed suffers as a result. Fortunately, there is a remedy for this ailment: {\em variance-reduction} (VR) \citep{johnson2013accelerating}. The purpose of VR mechanisms is to steer away from the naive gradient estimators. Instead, VR mechanisms iteratively construct and apply a gradient estimator whose variance would eventually vanish. This allows for larger learning rates to be used safely, which accelerates training. Among the early VR-empowered SGD methods belong SAG~\cite{roux2012stochastic}, SVRG~\cite{johnson2013accelerating}, SAGA~\cite{defazio2014saga}, and Finito~\cite{pmlr-v32-defazio14}. For a recent survey of VR methods, see \cite{gower2020variance}.

{\bf Random reshuffling.} While virtually all theoretical development concerning vanilla SGD and its VR-variants is devoted to {\em with-replacement} sampling procedures \cite{sigma_k}, widely-used machine learning software implements a different strategy instead: one based on {\em random reshuffling} (RR) of the training data~\cite{bottou2009curiously,recht2013parallel}. In RR methods, data points are randomly permuted and then sampled and processed in order, {\em without replacement}. This process is repeated several times until a model of suitable qualities is found. RR often outperforms SGD in practice~\cite{recht2013parallel}, and because of this, acts as the de-facto default sampling mechanism in deep learning~\cite{Bengio2012,Sun2020}.

{\bf Shuffle once or not at all.} RR does not live alone; it has cousins. {\em Shuffle-Once} (SO)~\cite{Nedic2001} differs from RR in that shuffling occurs only once, at the very beginning, before the training begins. The purpose of this procedure is to break the potentially adversarial default ordering of the data that could negatively affect training speed. It is also possible to decide not to worry about this and to process the data it the deterministic cyclic order it is readily available in. This idea is the basis of 
the {\em Cyclic GD} method~\cite{Luo1991, Grippo1994}.

{\bf Difficulties with analyzing RR and its cousins.} The main difficulty in analyzing this type of methods is that each gradient step within an epoch is {\em biased}, and performing a sharp analysis of methods based on biased estimators is notoriously difficult. While cyclic GD was studied already a few decades ago~\cite{Mangasarian1994,Bertsekas2000}, convergence rates were established relatively recently~\cite{Li2019,Ying2019,Gurbuzbalaban2019IG,Nguyen2020}. For the SO method, the situation is more complicated, and non-vacuous theoretical analyses were only performed recently~\cite{safran2020good, Rajput2020}. RR is well understood for twice-smooth~\cite{gurbuzbalaban2019random,haochen2018random} and smooth~\cite{Nagaraj2019} objectives. Moreover, lower bounds for RR and similar methods were also recently established~\cite{safran2020good, Rajput2020}. \citet{mishchenko2020random} performed an in-depth analysis of RR, SO and Cyclic GD with novel and simpler proof techniques, leading to improved and new convergence rates. Their rate for SO, for example, tightly matches the lower bound of \citet{safran2020good} in the case when each $f_i$ is strongly convex. However, despite these advances, RR and related method described above still suffer from the same problem as SGD, i.e., we do not have variants with linear convergence to the exact minimizer. Further, RR can be accelerated~\cite{gurbuzbalaban2019random}, and for small constant step-sizes, the neighborhood of solution can be controlled~\cite{sayed2014adaptation}.

Some cyclic and random reshuffling versions of variance-reduced methods were shown to obtain linear convergence. Incremental Average Gradient (IAG)---a cyclic version of the famous SAG method---was analyzed \cite{Gurbuzbalaban2017}. Based on this, the Double Incremental Average Gradient method was introduced, and it has a significantly better rate if each $f_i$ is strongly convex~\cite{mokhtari2018surpassing}. A linear rate for Cyclic SAGA was established by \citet{park2020linear}. The first analysis of Random Reshuffling with variance reduction was done by \citet{ying2020variance}. Firstly, they establish a linear rate for SAGA under random reshuffling, and then they introduce a new method called Amortized Variance-Reduced Gradient (AVRG), which is similar to SAGA. SVRG using RR was introduced by \citet{shamir2016without}, and their theoretical analysis was conducted for the Least Squares problem.

\section{Contributions}
%	While variance-reduced methods using sampling with replacement have been studied in detail in recent years, there has been very little work devoted to the analysis of variance-reduced methods under random reshuffling. In this paper we try to narrow this gap.
	
In this paper, we equip the Random Reshuffling, Shuffle-Once and Cyclic GD algorithms with a {\em variance reduction mechanism}. Our approach is based on an iterative reformulation of the finite sum problem \eqref{eq:main_finite_sum} via {\em controlled linear perturbations}. These reformulations are governed by an auxiliary sequence of control vectors, and their role is to obtain progressively improved conditioning of the problem. To the best of our knowledge, we provide the first convergence analysis of SVRG under random reshuffling (RR-SVRG). We also provide the first convergence analysis of SO-SVRG and Cyclic SVRG. Our theory leads to improved rates for variance-reduced algorithms under random reshuffling in all regimes considered, especially in the big data regime (Section~\ref{subsection 5}). We also provide a better rate for cyclic methods in the strongly convex and convex cases (Section~\ref{subsection 6}). 
%	\subsubsection{Convergence rates for RR-SVRG, SO-SVRG and Cyclic SVRG}

We provide theoretical guaranties in Section~\ref{section 5}. A summary of our complexity results as well as accounting for the required memory is presented in Table~\ref{Tab:mainresults}.

$\diamond$ \textbf{Strongly convex case.} If $f$ is strongly convex, we obtain $\mathcal{O}\left(\kappa^{3/2}\log \frac{1}{\varepsilon}\right)$ iteration (epoch-by-epoch) complexity for RR-SVRG, where $\kappa$ is the condition number. This rate is better than the $\mathcal{O}\left(\kappa^{2}\log \frac{1}{\varepsilon}\right)$ rate of RR-SAGA and AVRG introduced by~\citet{ying2020variance}. Moreover, if $n> \mathcal{O}(\kappa)$, we improve this rate for RR-SVRG and get $\mathcal{O}\left(\kappa\log \frac{1}{\varepsilon}\right)$ complexity. If each $f_i$ is strongly convex and the number of function is sufficiently large (Theorem~\ref{th3}), then the rate of RR-SVRG can be further improved to $\mathcal{O}\left(\kappa\sqrt{\frac{\kappa}{n}}\log \left(\frac{1}{\varepsilon}\right)\right)$. For Cyclic-SVRG we prove similar convergence results under the assumption of strong convexity of $f$. The iteration complexity of this method is $\mathcal{O}\left(\kappa^{3/2}\log \frac{1}{\varepsilon}\right)$, which is noticeably better than the $\mathcal{O}\left(n\kappa^2\log \frac{1}{\varepsilon}\right)$ rate of IAG~\cite{Gurbuzbalaban2017}. Furthermore, it is better than the $\mathcal{O}\left(\kappa^2\log\frac{1}{\varepsilon}\right)$ rate of Cyclic SAGA~\cite{park2020linear}. It is worth mentioning that \citet{mokhtari2018surpassing} obtain a better complexity, $\mathcal{O}\left(\kappa\log \frac{1}{\varepsilon}\right)$, for their DIAG method. However, their analysis requires much stricter assumption.

$\diamond$ \textbf{Convex case.} In the general convex setting we give the first analysis and convergence guarantees for RR-SVRG, SO-SVRG and Cyclic SVRG. After applying variance reduction, we obtain fast convergence to the exact solution. As expected, these methods have the sublinear rate $\mathcal{O}(\frac{1}{\varepsilon})$ in an ergodic sense.

$\diamond$ \textbf{Generalized version of RR-SVRG.} Finally, we introduce RR-VR -- a new RR-based method employing variance reduction ideas similar to those behind L-SVRG~\cite{kovalev2020don}. RR-VR can also be seen as a variant of RR-SVRG in which the control vectors are updated in a randomized manner, which allows for less frequent updates of the control vector.

\begin{algorithm}[t]
   \caption{RR-SVRG}
   \label{alg:RRSVRG}
\begin{algorithmic}
		\STATE \textbf{Input:} Stepsize $\gamma>0$, $y_0 = x_0 = x_0^0 \in \mathbb{R}^{d}$, number of epochs $T$.
		\FOR{$t =  0, 1, \dots T-1$ }
		\STATE {\color{blue}Sample a permutation $\{\pi_0, \dots, \pi_{n-1}\}$ of $\{1, \dots, n\}$} 
		\STATE $x_t^0 = x_t$
		\FOR{$i= 0, \dots, n-1$ }
		\STATE $g^i_t(x_t^i,y_t) =  \nabla f_{\pi_i} (x_t^i)-\nabla f_{\pi_i} (y_t)+\nabla f (y_t) $
		\STATE $x^{i+1}_t = x^i_t - \gamma g_t^i(x_t^i, y_t)$
		\ENDFOR
		\STATE $x_{t+1} = x^n_t$
		\STATE $y_{t+1} = x^n_t$
		\ENDFOR
\end{algorithmic}
\end{algorithm}

\begin{algorithm}[t]
   \caption{SO-SVRG}
   \label{alg:SOSVRG}
\begin{algorithmic}
		\STATE \textbf{Input:} Stepsize $\gamma>0$, $y_0 = x_0 = x_0^0 \in \mathbb{R}^{d}$, number of epochs $T$.
		\STATE {\color{blue}Sample a permutation $\{\pi_0, \dots, \pi_{n-1}\}$ of $\{1, \dots, n\}$} 		
		\FOR{$t =  0, 1, \dots T-1$ }
		\STATE $x_t^0 = x_t$
		\FOR{$i= 0, \dots, n-1$ }
		\STATE $g^i_t(x_t^i,y_t) =  \nabla f_{\pi_i} (x_t^i)-\nabla f_{\pi_i} (y_t)+\nabla f (y_t) $
		\STATE $x^{i+1}_t = x^i_t - \gamma g_t^i(x_t^i, y_t)$
		\ENDFOR
		\STATE $x_{t+1} = x^n_t$
		\STATE $y_{t+1} = x^n_t$
		\ENDFOR
\end{algorithmic}
\end{algorithm}

\begin{algorithm}[t]
   \caption{CYCLIC-SVRG}
   \label{alg:CYCLICSVRG}
\begin{algorithmic}
		\STATE \textbf{Input:} Stepsize $\gamma>0$, $y_0 = x_0 = x_0^0 \in \mathbb{R}^{d}$, number of epochs $T$.
		\FOR{$t =  0, 1, \dots T-1$ }
		\STATE $x_t^0 = x_t$
		\FOR{$i= 0, \dots, n-1$ }
		\STATE $g^i_t(x_t^i,y_t) =  \nabla f_{\pi_i} (x_t^i)-\nabla f_{\pi_i} (y_t)+\nabla f (y_t) $
		\STATE $x^{i+1}_t = x^i_t - \gamma g_t^i(x_t^i, y_t)$
		\ENDFOR
		\STATE $x_{t+1} = x^n_t$
		\STATE $y_{t+1} = x^n_t$
		\ENDFOR
\end{algorithmic}
\end{algorithm}

\section{Description of Algorithms}

We now describe the procedure of {\em sampling without replacement}. For a set of indices $1,2,\ldots,n$ corresponding to training data, our algorithms construct a {\em permutation} $\left\{\pi_{0}, \pi_{1}, \ldots, \pi_{n-1}\right\}$ randomly or deterministically. In other words, data shuffling occurs. If this process is repeated, we call this {\em reshuffling}.

\subsection{RR-SVRG, SO-SVRG, Cyclic SVRG}

 In our RR-SVRG method (Algorithm~\ref{alg:RRSVRG}), data permutations are sampled at the beginning of each epoch. We then proceed with $n$ steps of the form
$$x^{i+1}_t = x^i_t - \gamma g_t^i(x_t^i).$$
In each step we calculate the stochastic gradient estimator $g^i_t(x_t^i,y_t)$ using the current point $x^i_t$ and a control vector $y_t$:
$$g^i_t(x_t^i,y_t) =  \nabla f_{\pi_i} (x_t^i)-\nabla f_{\pi_i} (y_t)+\nabla f (y_t). $$
After each epoch we update the control vector $y_t$: $y_{t+1} = x^n_t,$ and
then start a new epoch. 

RR-SVRG is a version of SVRG, where a number of inner steps $m$ is equal to $n$, and in which sampling {\em without} replacement is used. \citet{johnson2013accelerating} remarked that $m = \mathcal{O}(n)$ works well in practice, but a theoretical analysis of this was not provided.

In SO-SVRG (Algorithm~\ref{alg:SOSVRG}) instead of doing data permutations in every iteration, we shuffle the data points randomly at the beginning only, and we use this one permutation in all subsequent epochs. The rest of the algorithm is the same as for RR-SVRG. 

Cyclic SVRG is equivalent to SO-SVRG, with one difference. The initial permutation is deterministic, or no permutation is performed at all and the data is processed in some natural order. The steps are then performed incrementally through all data, in the same order in each epoch.
\subsection{RR - VR}
\begin{algorithm}[t]
	\caption{Random Reshuffling with Variance Reduction}
	\label{alg:RR_VR}	
	\begin{algorithmic}[1]
		\STATE \textbf{Input:} Stepsize $\gamma>0$, probability $p$, $x_0 = x_0^0 \in \mathbb{R}^{d}, y_0 \in \mathbb{R}^{d}$, number of epochs $T$.
		\FOR{$t =  0, 1, \dots T-1$ }
		\STATE {\color{blue}Sample a permutation $\{\pi_0, \dots, \pi_{n-1}\}$ of $\{1, \dots, n\}$}
		\STATE $x_t^0 = x_t$
		\FOR{$i= 0, \dots, n-1$ }
		\STATE $g^i_t(x_t^i,y_t) =  \nabla f_{\pi_i} (x_t^i)-\nabla f_{\pi_i} (y_t)+\nabla f (y_t) $
		\STATE $x^{i+1}_t = x^i_t - \gamma g^i_t(x_t^i,y_t)$
		\ENDFOR
		\STATE $x_{t+1} = x^n_t$
		\STATE  $y_{t+1}=\begin{cases} y_t & \text{with probability } 1-p \\ x_t & \text{with probability } p \end{cases}$
		\ENDFOR
	\end{algorithmic}

\end{algorithm}
We also propose a generalized version of RR-SVRG, which we call RR-VR. The main idea of RR-VR (Algorithm~\ref{alg:RR_VR}) is that at the end of each epoch we flip a biased coin to decide whether to update the control vector $y_t$ or not. While in RR-SVRG the control vector $y_{t+1}$ is updated to the latest iterate $x_{t+1}$, in RR-VR we use the previous point $x_t$. We do this as it slightly simplified the analysis. However, it makes sense to use the newest point $x_{t+1}$ instead of $x_t$ to update the control vector in practice.

{\footnotesize
	
	\begin{table*}[t]
		\centering \footnotesize
		\caption{Comparison of the variance-reduced convergence results and implementations.}\label{table:1}
		\renewcommand{\arraystretch}{1.7}
		\begin{tabular}{|p{27mm}|p{27mm}|p{30mm}|p{12mm}|p{10mm}|p{35mm}|}
			
			\hline
			Algorithm &$\mu$-strongly convex $f_i$& $\mu$-strongly convex $f$ & convex $f$&memory& citation\\
			\hline
			\hline
			RR-SAGA
			& --
			&  $\mathcal{O}\left(\kappa^2\log \frac{1}{\epsilon}\right)$& -- &$\mathcal{O}(dn)$&\citet{ying2020variance}\\ \hline
			
			AVRG
			& --
			& $\mathcal{O}\left(\kappa^2\log \frac{1}{\epsilon}\right)$&--& $\mathcal{O}(d)$&\citet{ying2020variance}\\ \hline
			\rowcolor{bgcolor}
			RR/SO-SVRG& $\mathcal{O}\left(\kappa\sqrt{\frac{\kappa}{n}} \log \frac{1}{\epsilon}\right)$ \footnotesize(in~Big~Data~regime)&$\mathcal{O}\left( \kappa \log \frac{1}{\epsilon} \right)$ \footnotesize(in~Big~Data~regime) $\mathcal{O}\left( \kappa\sqrt{\kappa} \log \frac{1}{\epsilon} \right)$      ~~~~~\footnotesize(in~general~regime)&$\mathcal{O}\left(\frac{L}{\varepsilon}\right)$&$\mathcal{O}(d)$& this paper\\
			\hline
			Cyclic SAGA& $\mathcal{O}\left(\kappa^2\log \frac{1}{\epsilon}\right)$&--&--&$\mathcal{O}(dn)$&\citet{park2020linear}\\
			\hline
			IAG (Cyclic SAG)& -- & $\mathcal{O}\left(n\kappa^2\log \frac{1}{\epsilon}\right)$&--&$\mathcal{O}(dn)$&\citet{Gurbuzbalaban2017}\\
			\hline
			DIAG (Cyclic Finito)&  $\mathcal{O}\left(\kappa\log \frac{1}{\epsilon}\right)$&--&--&$\mathcal{O}(dn)$&\citet{mokhtari2018surpassing}\\
			\hline
			\rowcolor{bgcolor}
			Cyclic SVRG&--&$\mathcal{O}\left(\kappa\sqrt{\kappa}\log \frac{1}{\epsilon}\right)$&$\mathcal{O}\left(\frac{L}{\varepsilon}\right)$&$\mathcal{O}(d)$& this paper\\
			\hline
		\end{tabular}
		\label{Tab:mainresults}

	\end{table*}
}

\section{Assumptions and Notation}
Before introducing our convergence results, let us first formulate the definitions and assumptions we use throughout the work.  Function $f: \mathbb{R}^d \rightarrow \mathbb{R}$ is $L$-smooth if 
	\begin{equation}
	\squeeze
f(y) \leq f(x)+\left<\nabla f(x),y-x\right> +\frac{L}{2}\|y-x\|^{2}, \forall x, y\in \R^d,
	\end{equation}
convex if 
\begin{equation}
f(y) \geq f(x)+\left<\nabla f(x),y-x\right> \quad \forall x,y\in \R^d,
\end{equation}
and $\mu$-strongly convex if 
	\begin{equation}
	\squeeze 
	f(y) \geq f(x)+\left<\nabla f(x),y-x\right> + \frac{\mu}{2}\|y-x\|^2 \quad \forall x,y\in \R^d.
	\end{equation}

	The Bregman divergence with respect to $f$ is the mapping $D_f:\R^d\times \R^d\to \R$ defined as follows:
	\begin{equation}
	D_{f}(x, y) \eqdef f(x)-f(y)-\langle\nabla f(y), x-y\rangle.
	\end{equation}
	Note that if $y=x_*$, where $x^*$ is a minimum of $f$, then we have 
	\begin{equation*}
	D_{f}(x, x_*) = f(x) - f(x_*).
	\end{equation*}

Lastly, we define an object that plays the key role in our analysis. 
\begin{definition}[Variance at optimum]
	\label{def:sigma}
	Gradient variance at optimum is the quantity
	\begin{equation}
	\squeeze
	\sigma_{*}^{2} \eqdef \frac{1}{n} \sum \limits_{i=1}^{n}\left\|\nabla f_{i}\left(x_{*}\right)\right\|^{2}.
	\end{equation}
\end{definition}
This quantity is used in several recent papers on stochastic gradient-type methods. Particularly, it is a version of gradient noise introduced in \citet{gower2019sgd} for finite sum problems.

For all theorems in this paper the following assumption is used. 
\begin{assumption}
	\label{L-smooth}
	The objective $f$ and the individual losses $f_1, \ldots , f_n$ are all $L$-smooth. We also assume the existence of a minimizer $x_* \in \mathbb{R}^d$. 
\end{assumption}
This assumption is classical in the literature, and it is necessary for us to get convergence results for all the methods described above.

\section{Variance Reduction via Controlled Linear Perturbations}
In the design of our methods we employ a simple but powerful tool: the idea of introducing a sequence of carefully crafted reformulations of the original finite sum problem, and applying RR on these instead of the original formulation. As the sequence is designed to have progressively better conditioning properties, RR will behave progressively better as well, and this is why the result is a variance reduced RR method. The main idea is to perturb the objective function with zero written as the average of $n$ nonzero linear functions. This perturbation is performed at the beginning of each epoch and stays fixed within each epoch. Let us consider the finite sum problem \eqref{eq:main_finite_sum} and vectors $a_i, \dots, a_n \in \R^d$ summing up to zero: $\sum_{i=1}^{n}a_i = 0$. Adding this zero to $f$, we reformulate problem~\ref{eq:main_finite_sum} into the equivalent form
\begin{align}
\label{reform}
f(x) &\eqdef  \squeeze \frac{1}{n}  \sum \limits_{i=1}^n(f_i(x)+\left\langle a_i,x \right\rangle) = \sum \limits_{i=1}^{n}\tilde{f}_i(x),
\end{align}
where $\tilde{f}_i(x) = f_i(x)+\left\langle a_i,x \right\rangle$. We should note that $\nabla\tilde{f}_i(x) = \nabla f_i(x)+ a_i$. Next, we establish a simple but important property of this reformulation.
\begin{proposition}
	\label{prop-reform}
	Assume that each $f_i$ is $\mu$-strongly convex (convex) and $L$-smooth. Then $\tilde{f}$ defined as 
	\begin{equation}
	\tilde{f}_i(x) = f_i(x)+\left\langle a_i,x \right\rangle
	\end{equation}
	is $\mu$-strongly convex (convex) and $L$-smooth.
\end{proposition}

In RR-SVRG we utilize the following gradient estimate: 
$$g^i_t(x_t^i,y_t) =  \nabla f_{\pi_i} (x_t^i)-\nabla f_{\pi_i} (y_t)+\nabla f (y_t). $$
This estimate arises from the above reformulated. Indeed, $g^i_t(x_t^i,y_t) =  \nabla f_{\pi_i} (x_t^i)+a_i,$
where $a_i = -\nabla f_{\pi_i} (y_t)+\nabla f (y_t).$ Trivially, the sum of these vectors is equal to zero:
\[\squeeze \sum \limits_{i=1}^{n} a_i = -\sum \limits_{i=1}^{n}\nabla f_{\pi_i} (y_t)+\sum \limits_{i=1}^{n}\nabla f (y_t)
= 0.
\]
The idea behind this approach is simple. Since the reformulated problem satisfies all assumptions of the original problem, we can apply theorems of methods that are not variance-reduced. Then, by updating the control vector, we can get an upper bound for the variance. The critical requirement for variance reduction mechanism is updating the control vector. We update it after each epoch, which means that the problem's reformulation happens at the beginning of the next epoch. However, RR-VR (Algorithm~\ref{alg:RR_VR}) allows to do this probabilistically.

Now we are ready to formulate the core lemma of our work. This lemma is simple but it allows us to make algorithms variance-reduced.
\begin{lemma}
	\label{main_lemma_lemma}
	Assume that each $f_i$ is $L$-smooth and convex. If we apply the linear perturbation reformulation~\eqref{reform} 
	using vectors of the form $ a_i = -\nabla f_{\pi_i} (y_t)+\nabla f (y_t)$, then the variance of reformulated problem satisfies the following inequality:
	\begin{equation}
	\label{main_lemma}\squeeze 
	\tilde{\sigma}_{*}^{2}=\frac{1}{n} \sum \limits_{i=1}^{n}\left\|\nabla \tilde{f}_{i}\left(x_{*}\right)\right\|^{2}\leq 4L^2\|y_t-x_*\|^2.
	\end{equation}
\end{lemma}

\section{Convergence Analysis}\label{section 5}
Having described the methods and the idea of controlled linear perturbations, we are ready to proceed to the formal statement of our convergence results.

\subsection{Convergence Analysis of RR-SVRG and SO-SVRG}\label{subsection 5}

\subsubsection{Strongly Convex Objectives}
We provide two different rates in the strongly convex case. 
\begin{theorem} 
	\label{th1}
	Suppose that each $f_i$ is convex, $f$ is $\mu$-strongly convex, and Assumption~\ref{L-smooth}
	holds. Then provided the stepsize satisfies $\gamma \leq \frac{1}{2\sqrt{2} L n}\sqrt{\frac{\mu}{L}},$
	the iterates generated by RR-SVRG (Algorithm~\ref{alg:RRSVRG}) or by SO-SVRG (Algorithm~\ref{alg:SOSVRG}) satisfy
	\begin{align*}
	\squeeze 
	\mathbb{E} \left[ \|x_T - x_* \|^2 \right] \leq \left( 1 - \frac{\gamma n \mu}{2} \right)^T \|x_0 - x_*\|^2.
	\end{align*}
\end{theorem}
\begin{corollary}
	\label{corollary1}
	Suppose that the assumptions in Theorem~\ref{th1} hold. Then the iteration complexity of Algorithms~\ref{alg:RRSVRG} and~\ref{alg:SOSVRG} is
	\begin{equation*}
	\squeeze 
	T = \mathcal{O}\left(\kappa\sqrt{\kappa}\log \left(\frac{1}{\varepsilon}\right)\right).
	\end{equation*}
\end{corollary}
If we additionally assume that $n>\mathcal{O}(\kappa)$, then we can use a larger step-size, which leads to an improved rate.
\begin{theorem}
	\label{th2}
	Suppose that each $f_i$ is convex, $f$ is $\mu$-strongly convex and Assumption~\ref{L-smooth} holds. Additionally assume we are in the ``big data'' regime characterized by $n \geq \frac{2 L}{\mu} \cdot \frac{1}{1-\frac{\mu}{\sqrt{2} L}}$. Then provided the stepsize satisfies $\gamma \leq \frac{1}{\sqrt{2}Ln},$
	the iterates generated by RR-SVRG (Algorithm~\ref{alg:RRSVRG}) or by SO-SVRG (Algorithm~\ref{alg:SOSVRG}) satisfy
	\begin{align*}
	\squeeze 
	\mathbb{E} \left[ \|x_T - x_* \|^2 \right] \leq \left( 1 - \frac{\gamma n \mu}{2} \right)^T \|x_0 - x_*\|^2.
	\end{align*}
\end{theorem}
This additional assumption allowed us to make a significant improvement in the iteration complexity.
\begin{corollary}
	\label{corollary2}
	Suppose that assumptions in Theorem~\ref{th2} hold. Then the iteration complexity of Algorithms~\ref{alg:RRSVRG} and~\ref{alg:SOSVRG} is
	\begin{equation*}
	\squeeze 
	T = \mathcal{O}\left(\kappa\log \left(\frac{1}{\varepsilon}\right)\right).
	\end{equation*}
\end{corollary}
As we shall now see, we have an even better rate in the case when each function $f_i$ is strongly convex. 

\begin{theorem}\label{th3}
	Suppose that the functions $f_1, \ldots, f_n$ are $\mu$-strongly convex and Assumption~\ref{L-smooth} holds. Fix constant $0<\delta<1$. If the stepsize satisfies $\gamma\leq\frac{\delta}{L}\sqrt{\frac{\mu}{2nL}}$ and if number of functions is sufficiently big, $n>\log\left(\frac{1}{1-\delta^2}\right)\cdot\left(\log\left(\frac{1}{1-\gamma\mu}\right)\right)^{-1}$, then the iterates generated by RR-SVRG (Algorithm~\ref{alg:RRSVRG}) or by SO-SVRG (Algorithm~\ref{alg:SOSVRG}) satisfy
	\begin{align*}
	\squeeze 
	\mathbb{E} \left[ \|x_T - x_* \|^2 \right] \leq \left( \left(1 - \gamma \mu\right)^n +\delta^2 \right)^T \|x_0 - x_*\|^2.
	\end{align*}
\end{theorem}
We need to use an additional assumption to get a better convergence rate. 
\begin{corollary}
	\label{corollary3}
	Suppose that assumptions in Theorem~\ref{th3} hold. Additionally assume that $n$ is large enough to satisfy the inequality $\delta^2 \leq (1-\gamma\mu)^{\frac{n}{2}}\left(1-(1-\gamma\mu)^{\frac{n}{2}}\right)$. Then the iteration complexity of Algorithms~\ref{alg:RRSVRG} and~\ref{alg:SOSVRG} is 
	\begin{equation*}
	\squeeze 
	T = \mathcal{O}\left(\kappa\sqrt{\frac{\kappa}{n}}\log \left(\frac{1}{\varepsilon}\right)\right).
	\end{equation*}
\end{corollary}
This additional assumption is quite complicated, and relaxations could be further explored.

\subsubsection{Convex Objectives}
In our work we provide the first bounds for SVRG under random reshuffling without strong convexity.
\begin{theorem}
	\label{th4}
	Suppose the functions $f_1, f_2, \ldots, f_n$ are convex and Assumption~\ref{L-smooth} holds. Then for RR-SVRG (Algorithm~\ref{alg:RRSVRG}) or SO-SVRG (Algorithm~\ref{alg:SOSVRG}) with stepsize $\gamma \leq \frac{1}{\sqrt{2}Ln},$ the average iterate $\hat{x}_{T} \eqdef \frac{1}{T} \sum_{t=1}^{T} x_{t}$ satisfies 
	\begin{align*}
	\squeeze 
	\mathbb{E}\left[f\left(\hat{x}_{T}\right)-f\left(x_{*}\right)\right] \leq \frac{3\left\|x_{0}-x_{*}\right\|^{2}}{2 \gamma n T}.
	\end{align*}
\end{theorem}
\begin{corollary}
	\label{corollary4}
	Suppose that the assumptions in Theorem~\ref{th4} hold. Then the iteration complexity of algorithms is
	\begin{equation*}
	\squeeze 
	T = \mathcal{O}\left(\frac{L\left\|x_{0}-x_{*}\right\|^{2}}{\varepsilon}\right).
	\end{equation*}
\end{corollary}

\subsection{Convergence Analysis of Cyclic SVRG}\label{subsection 6}
In this section we present results for Cyclic SVRG. They are very similar to the previous bounds. However, the lack of randomization does not allow us to improve convergence in the big data regime.
\subsubsection{Strongly convex objectives}
\begin{theorem}
	\label{th5}
	Suppose that each $f_i$ is convex function, $f$ is $\mu$-strongly convex function, and Assumption~\ref{L-smooth}
	holds. Then provided the stepsize satisfies $\gamma \leq \frac{1}{4 L n}\sqrt{\frac{\mu}{L}},$
	the iterates generated by Cyclic SVRG (Algorithm~\ref{alg:CYCLICSVRG}) satisfy
	\begin{align*}
	\squeeze 
	\mathbb{E} \left[ \|x_T - x_* \|^2 \right] \leq \left( 1 - \frac{\gamma n \mu}{2} \right)^T \|x_0 - x_*\|^2.
	\end{align*}
\end{theorem}
This leads to the same complexity as that of RR/SO-SVRG.
\begin{corollary}
	\label{corollary5}
	Suppose the assumptions in Theorem~\ref{th5} hold. Then the iteration complexity is
	\begin{align*}
	\squeeze 
	T = \mathcal{O}\left(\kappa\sqrt{\kappa}\log\left(\frac{1}{\varepsilon}\right)\right).
	\end{align*}
\end{corollary}
Our rate for Cyclic SVRG is better than the rate of Cyclic SAGA~\cite{park2020linear}. We remark that the convergence rate of DIAG~\cite{mokhtari2018surpassing} is better still; however, their result requires strong convexity of each $f_i$. 

\subsubsection{Convex objectives}
Similarly, we can establish convergence results for Cyclic SVRG in the convex case. 
\begin{theorem}
	\label{th6}
	Suppose the functions $f_1, f_2, \ldots, f_n$ are convex and Assumption~\ref{L-smooth} hold.s Then for Algorithm~\ref{alg:CYCLICSVRG} with a stepsize $\gamma \leq \frac{1}{{2\sqrt{2}Ln}}$, the average iterate $\hat{x}_{T} \eqdef \frac{1}{T} \sum_{j=1}^{T} x_{j}$ satisfies 
	\begin{align*}
	\squeeze 
	\mathbb{E}\left[f\left(\hat{x}_{T}\right)-f\left(x_{*}\right)\right] \leq \frac{2\left\|x_{0}-x_{*}\right\|^{2}}{ \gamma n T}.
	\end{align*}
\end{theorem}
The complexity of Cyclic SVRG is equivalent to the complexity of RR/SO-SVRG up to a constant factor. 
\begin{corollary}
	\label{corollary6}
	Let the assumptions in the Theorem~\ref{th6} hold. Then the iteration complexity of Algorithm~\ref{alg:CYCLICSVRG} is
	\begin{equation*}
	\squeeze 
	T = \mathcal{O}\left(\frac{L\left\|x_{0}-x_{*}\right\|^{2}}{\varepsilon}\right).
	\end{equation*}
\end{corollary}

\begin{figure*}[th!]
	\centering
	\begin{tabular}{ccc}
		\includegraphics[scale=0.35]{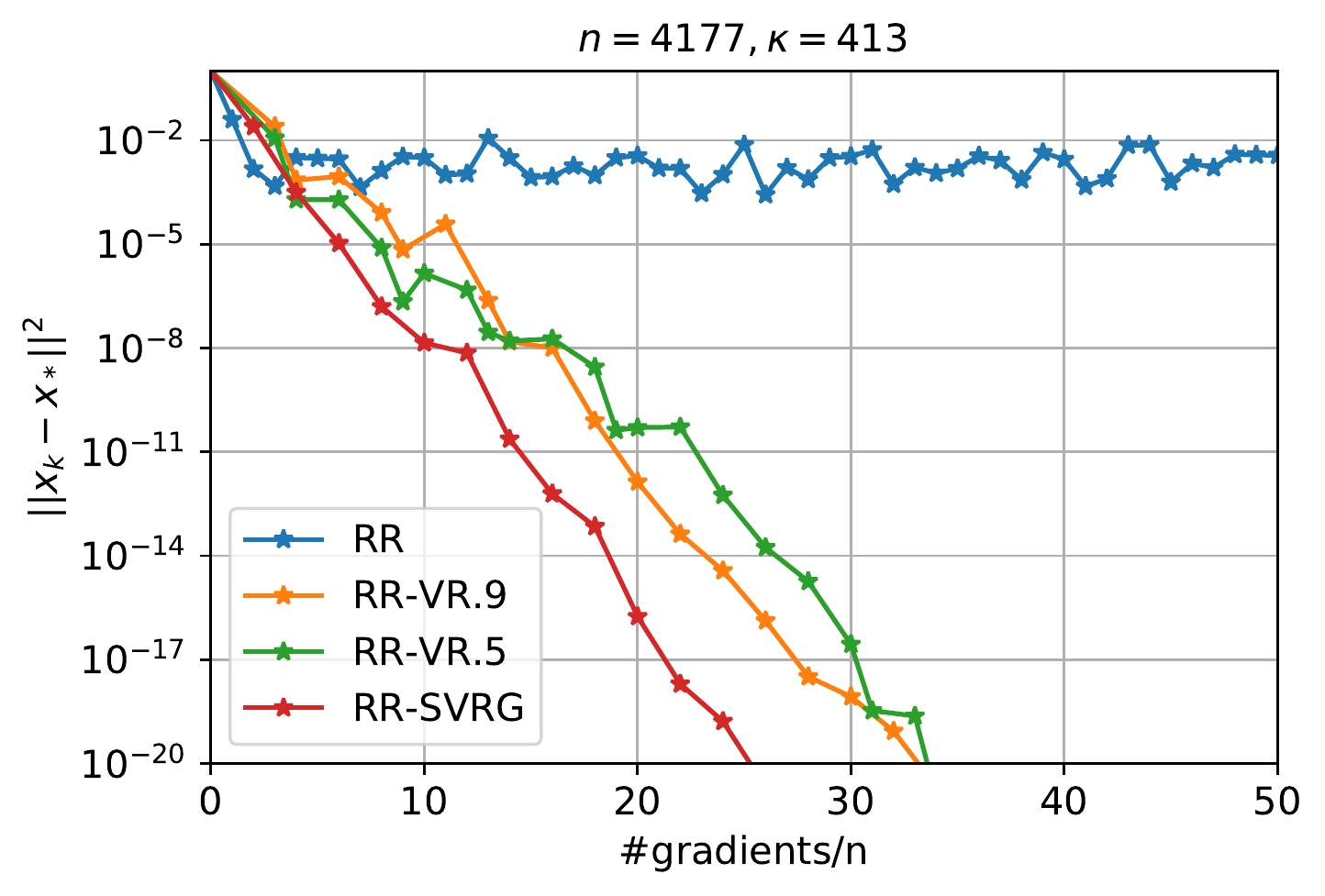}&
		\includegraphics[scale=0.35]{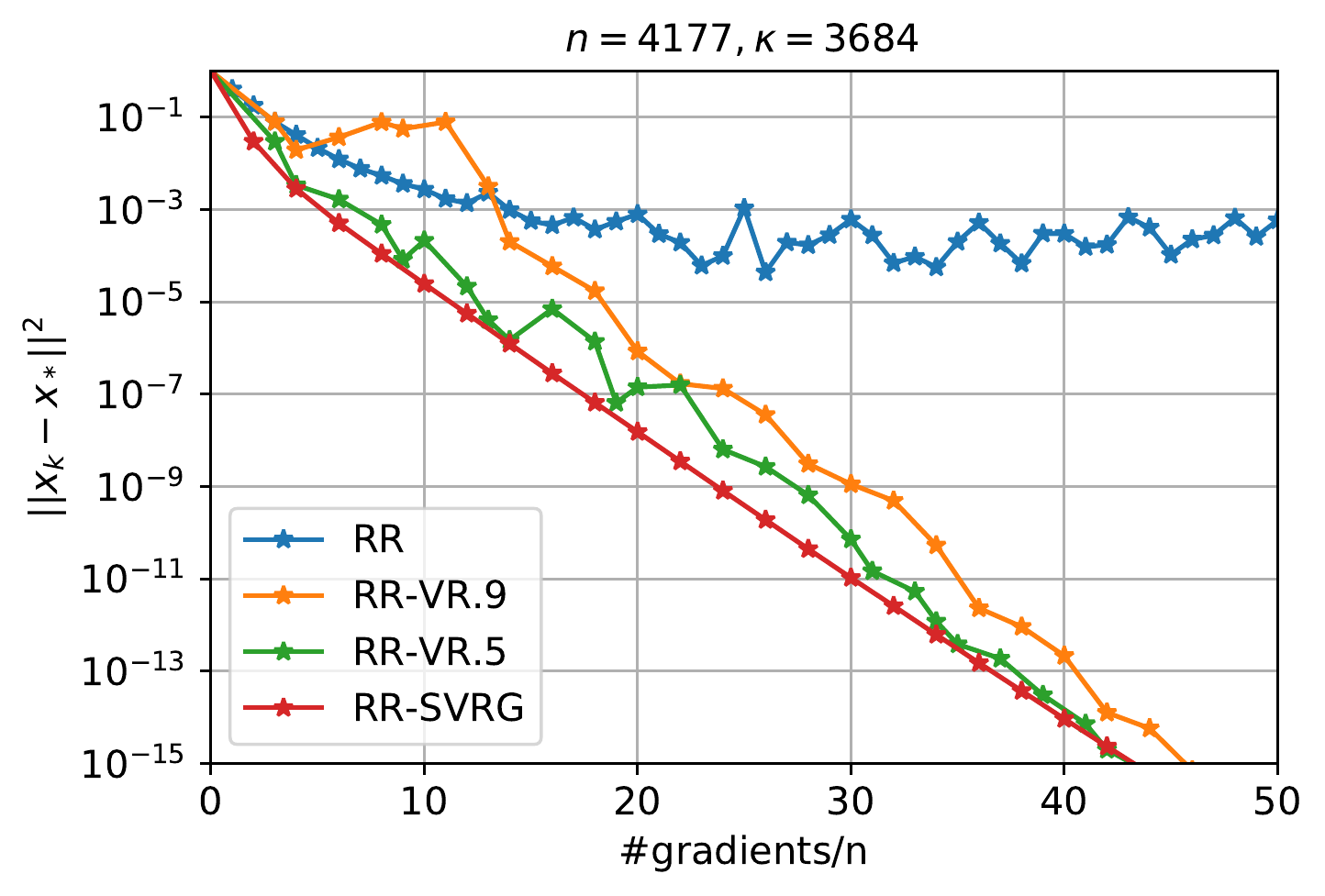}&
		\includegraphics[scale=0.35]{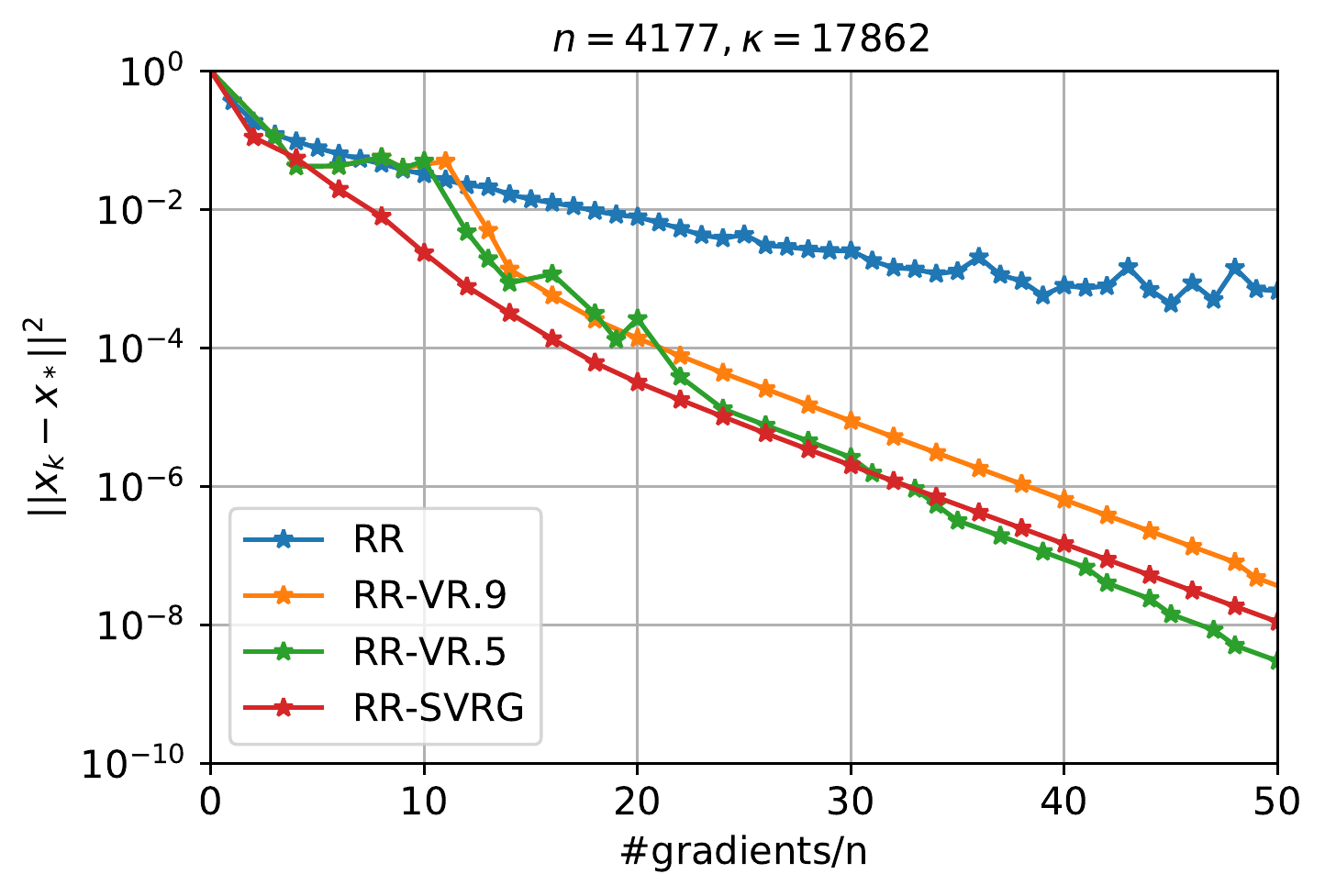}
		% 			(a)&(b)&(c)
	\end{tabular}
	\caption{Comparison of RR, RR-VR with probabilities $p=1$ (RR-SVRG), $p=0.5$ (RR-VR.5) and $p=0.9$ (RR-VR.9) on \texttt{abalone} dataset, We set the regularization constant $\lambda$ as $\frac{1}{n}, \frac{1}{10n}, \frac{10}{n}$ to obtain different condition numbers $\kappa=\frac{L}{\mu}$}.
	\label{fig:rr_vs_rrvr}
\end{figure*}

\begin{figure*}[th!]
	\centering
	\begin{tabular}{ccc}
		\includegraphics[scale=0.35]{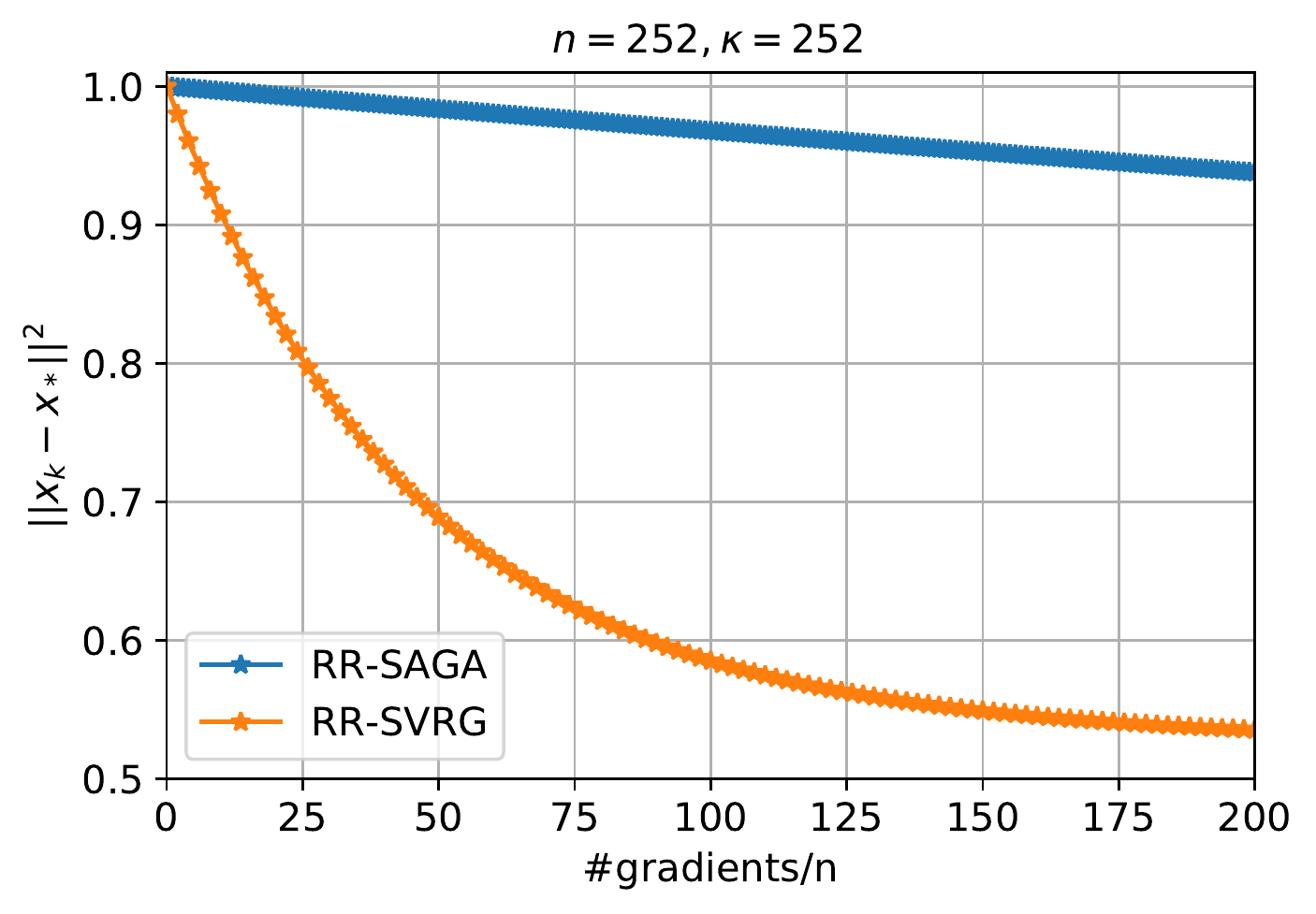}&
		\includegraphics[scale=0.35]{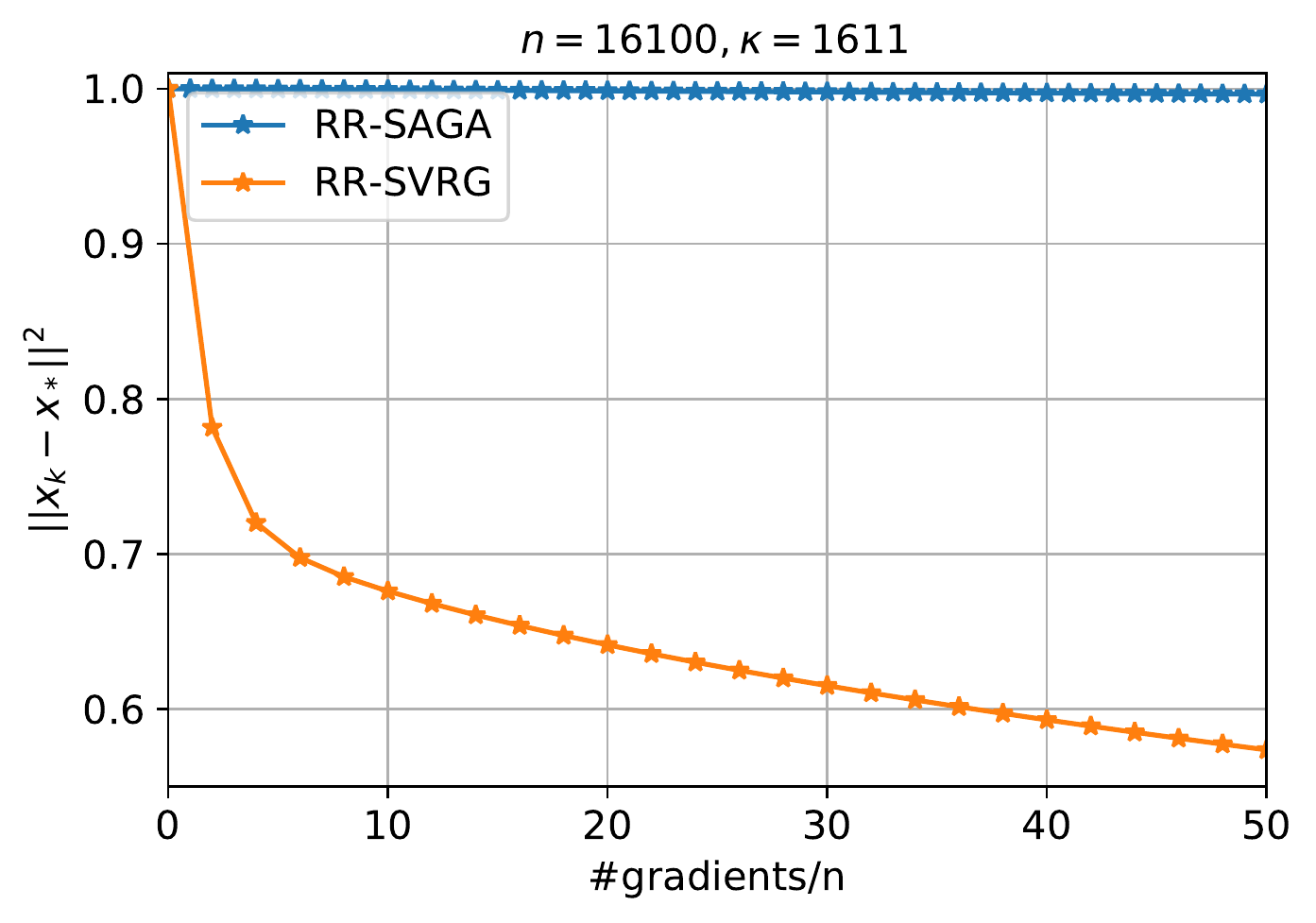} &
		\includegraphics[scale=0.35]{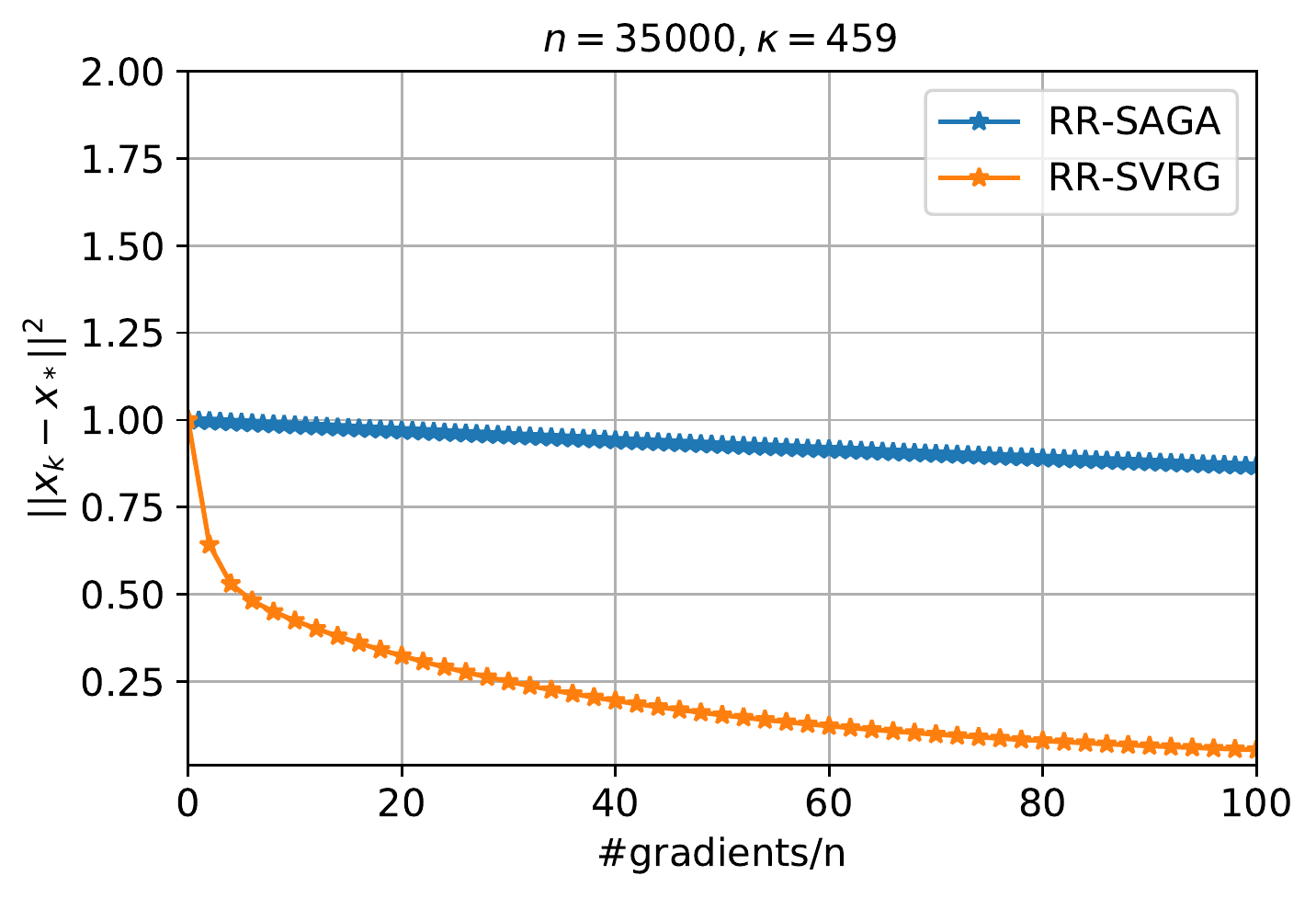}
	\end{tabular}
	\caption{Comparison of RR-SVRG and RR-SAGA with theoretical stepsizes on \texttt{bodyfat}, \texttt{a7a}, and \texttt{ijcnn1} datasets (from left to right). }
	\label{fig:rr_vs_rrvr}
\end{figure*}		

\subsection{Convergence Analysis of RR-VR}

In this section we formulate convergence results for a generalized version of SVRG under random reshuffling. Analysis of RR-VR (Algorithm~\ref{alg:RR_VR}) is more complicated. To analyze this method, we introduce Lyapunov functions. 
\begin{theorem}
	\label{th7}
	Suppose that each $f_i$ is convex, $f$ is $\mu$-strongly convex, and Assumption~\ref{L-smooth}
	holds. Then provided the parameters satisfy $n>\kappa$, $\frac{\kappa}{n}<p<1$ and $\gamma \leq \frac{1}{2\sqrt{2}Ln}$, 
	the final iterate generated by RR-VR (Algorithm~\ref{alg:RR_VR}) satisfies
	\begin{align*}
	V_{T} \leq \max \left( q_1,q_2 \right)^{T} V_{0},
	\end{align*}
	where
	\begin{align*}
	\squeeze 
	q_1 = 1-\frac{\gamma \mu n}{4}\left(1-\frac{p}{2}\right), \quad
	q_2 = 1-p+\frac{8}{\mu} \gamma^{2} L^{3} n,
	\end{align*}
	and the Lyapunov function is defined via
		\begin{align*}
	\squeeze 
	V_t \eqdef \mathbb{E}\left[\left\|x_{t}-x_{*}\right\|^{2}\right]+\left(\frac{4}{\gamma\mu n}\right)^{-1}\mathbb{E}\left[\left\|y_{t}-x_{*}\right\|^{2}\right]. 
	\end{align*}
\end{theorem}
Note that the probability $p$ should not be too small. 
\begin{corollary}
	\label{corollary7}
	Suppose the assumptions in Theorem~\ref{th7} hold. Then the iteration complexity of Algorithm~\ref{alg:RR_VR} is
	\begin{align*}
\squeeze	
T = \mathcal{O}\left(\kappa\log \left(\frac{1}{\varepsilon}\right)\right).
	\end{align*}
\end{corollary}
We obtain the same complexity as that of of RR-SVRG. 

\begin{theorem}
	\label{th_last}
	Suppose that the functions $f_1, \ldots, f_n$ are $\mu$-strongly convex, and that Assumption~\ref{L-smooth} holds. Then for RR-VR (Algorithm~\ref{alg:RR_VR}) with parameters that satisfy $\gamma \leq \frac{1}{2L}\sqrt{\frac{\mu}{2nL}}$, $\frac{1}{2}<\delta<\frac{1}{\sqrt{2}}$, $0<p<1$, and for a sufficiently large number of functions, $n>\log\left(\frac{1}{1-\delta^2}\right)\cdot\left(\log\left(\frac{1}{1-\gamma\mu}\right)\right)^{-1}$, the iterates generated by the RR-VR algorithm satisfy
	\begin{align*}	
	V_{T} &\squeeze	
\leq \max \left(q_1,q_2\right)^{T} V_{0},	
	\end{align*}
	where $$ \squeeze q_1 
= (1-\gamma \mu)^{n}+\delta^2, \quad
	q_2 	
= 1-p\left(1-\frac{2\gamma^2L^3n}{\mu\delta^2}\right),$$ and 
$$
	\squeeze
	V_t \eqdef \mathbb{E}\left[\left\|x_{t}-x_{*}\right\|^{2}\right]+\frac{\delta^2}{p}\mathbb{E}\left[\left\|y_{t}-x_{*}\right\|^{2}\right]. 
$$
\end{theorem}
\begin{corollary}
	\label{corollary8}
	Suppose the assumptions in Theorem~\ref{th_last} hold. Then iteration complexity of Algorithm~\ref{alg:RR_VR} is
	\begin{align*}
	\squeeze	
	T = \mathcal{O}\left(\max\left(\kappa\sqrt{\frac{\kappa}{n}},\frac{1}{2\log (2\delta)}\right)\log \left(\frac{1}{\varepsilon}\right)\right).
	\end{align*}
\end{corollary}
We get almost the same rate as the rate of RR-SVRG, but there is one difference. Complexity depends on $\delta$ term. However, the first term dominates in most cases. 

\section{Experiments}
	
\begin{figure*}[th!]
		\centering
		\begin{tabular}{ccc}
\includegraphics[scale=0.35]{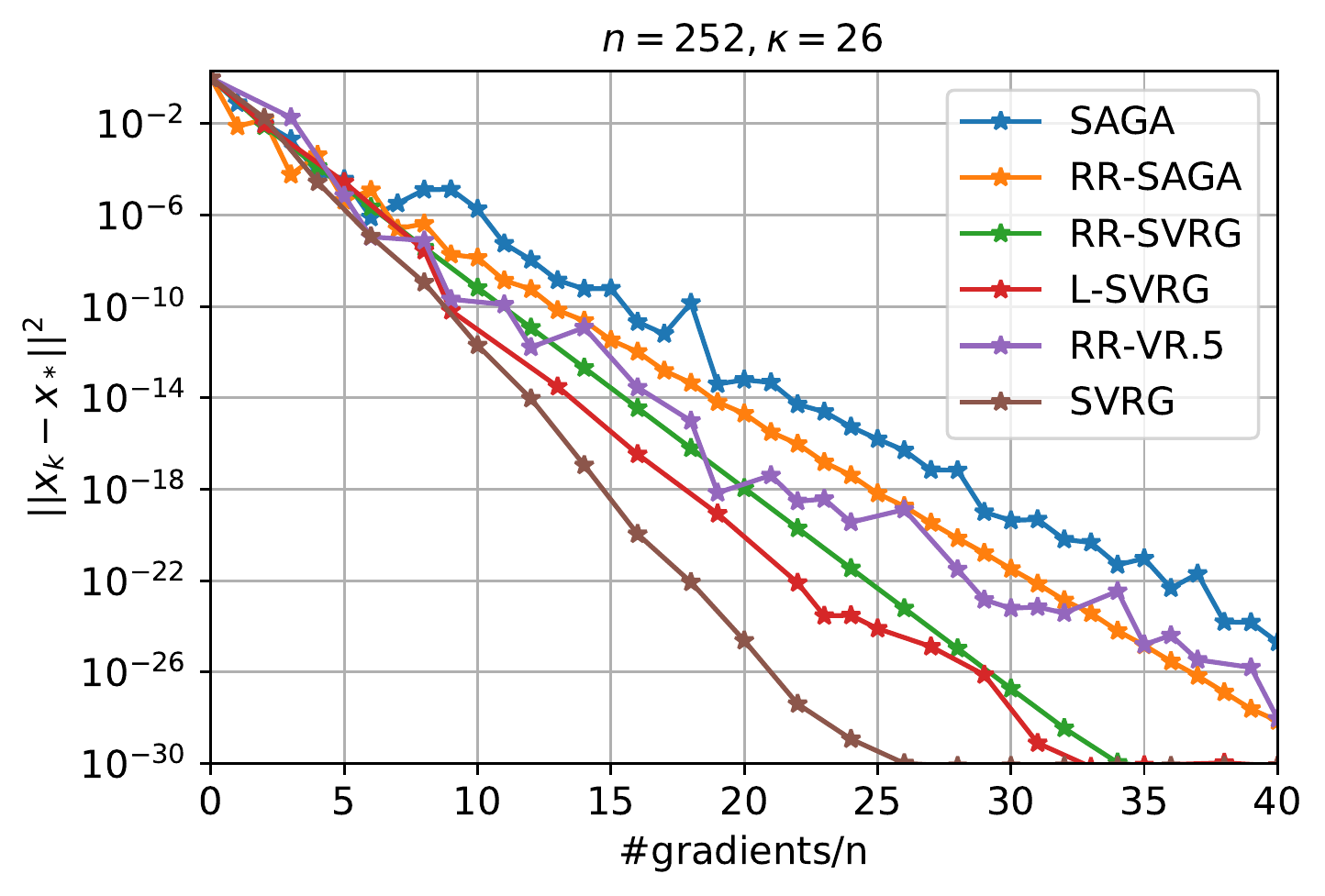} &
	\includegraphics[scale=0.35]{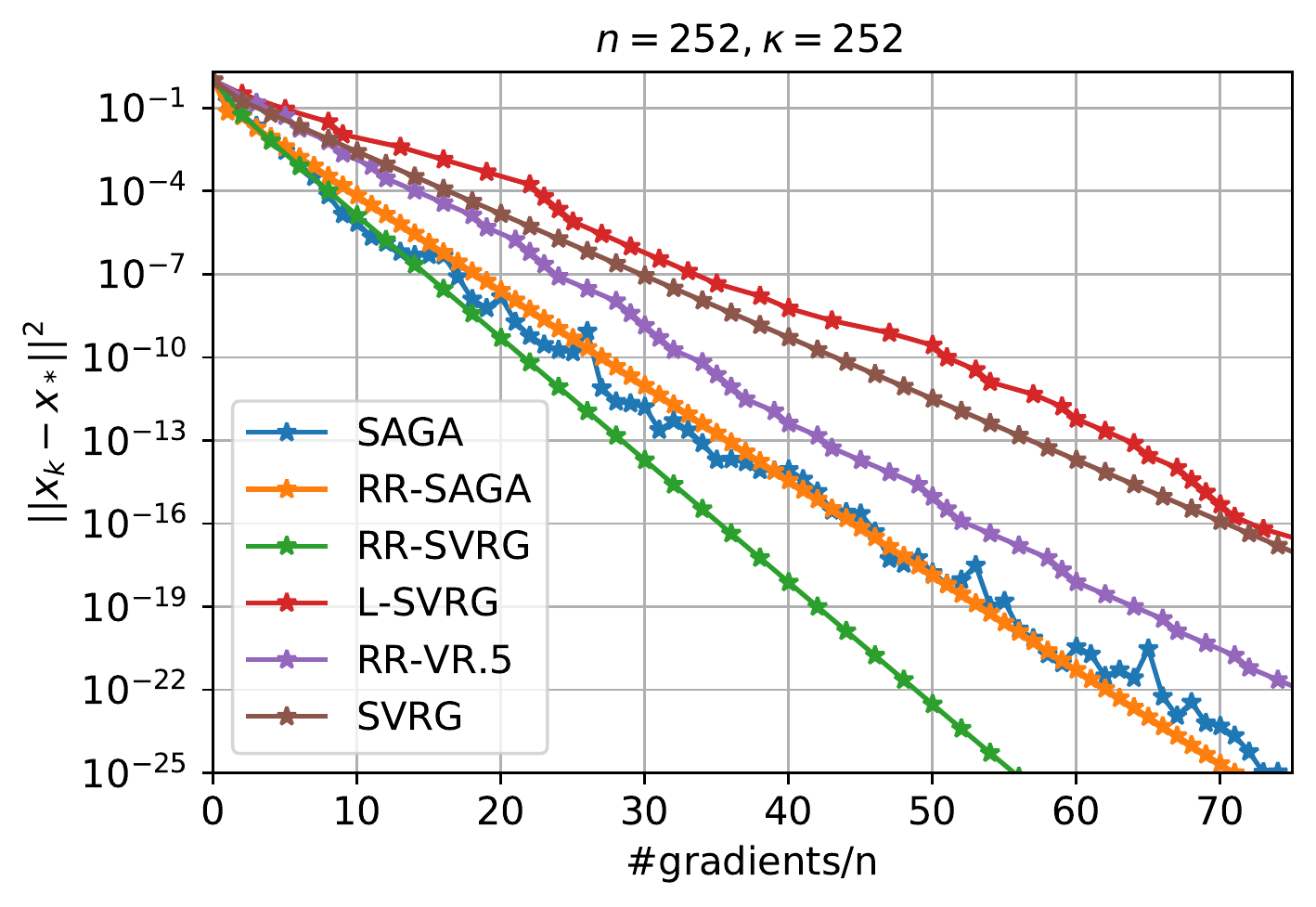} &
	
 \includegraphics[scale=0.35]{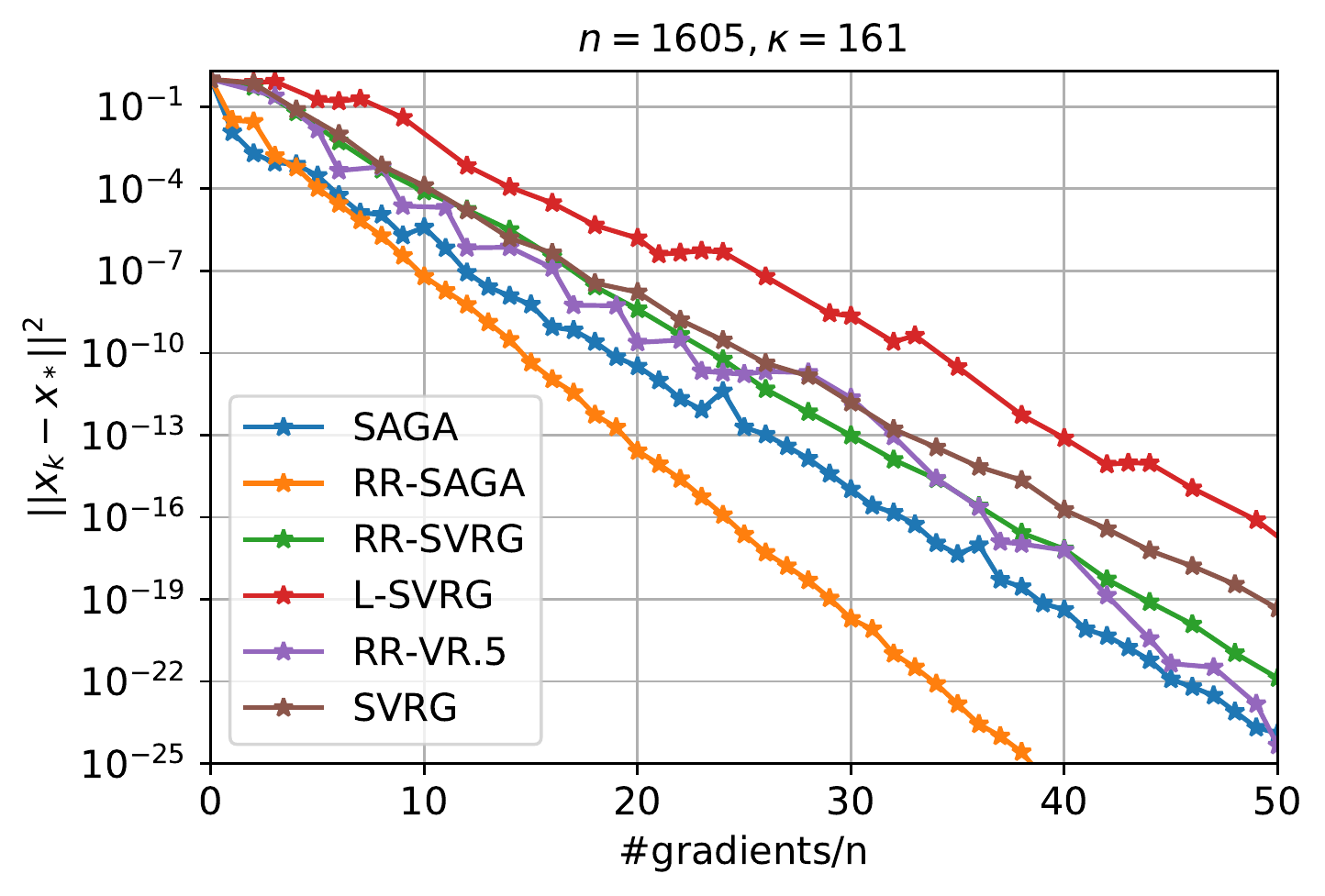} 

\end{tabular}
\caption{Comparison of SAGA, RR-SAGA, RR-SVRG, L-SVRG, SVRG and RR-VR with $p=0.5$ (RR-VR.5) with optimal stepsizes on \texttt{bodyfat} dataset with different regularization constants (on the left and middle) and \texttt{a1a} (on the right).}
\label{fig:saga_svrg_rrvr}
\end{figure*}

\begin{figure*}[th!]
	\centering
		\begin{tabular}{cccc}
			\includegraphics[scale=0.26]{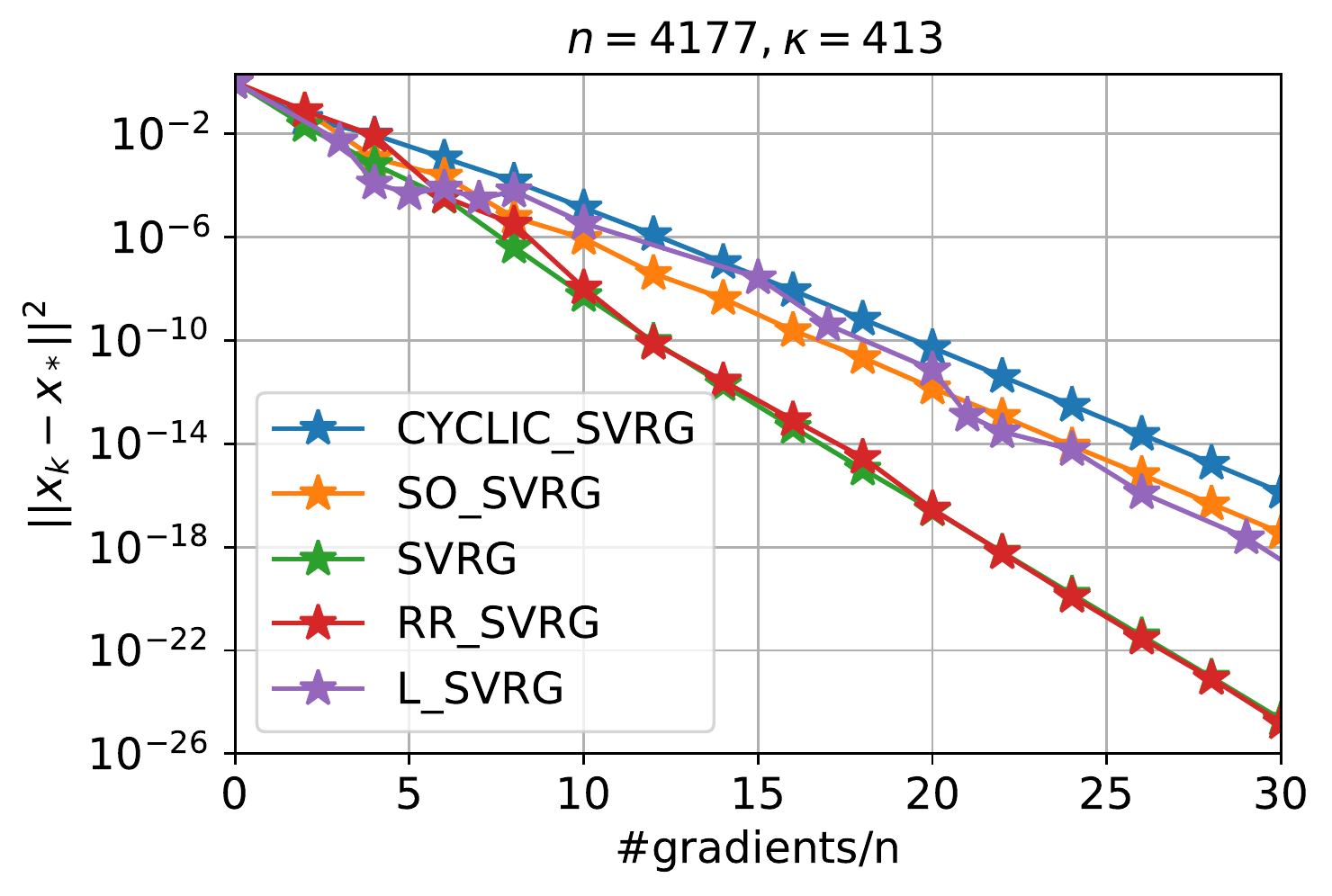}&
			\includegraphics[scale=0.26]{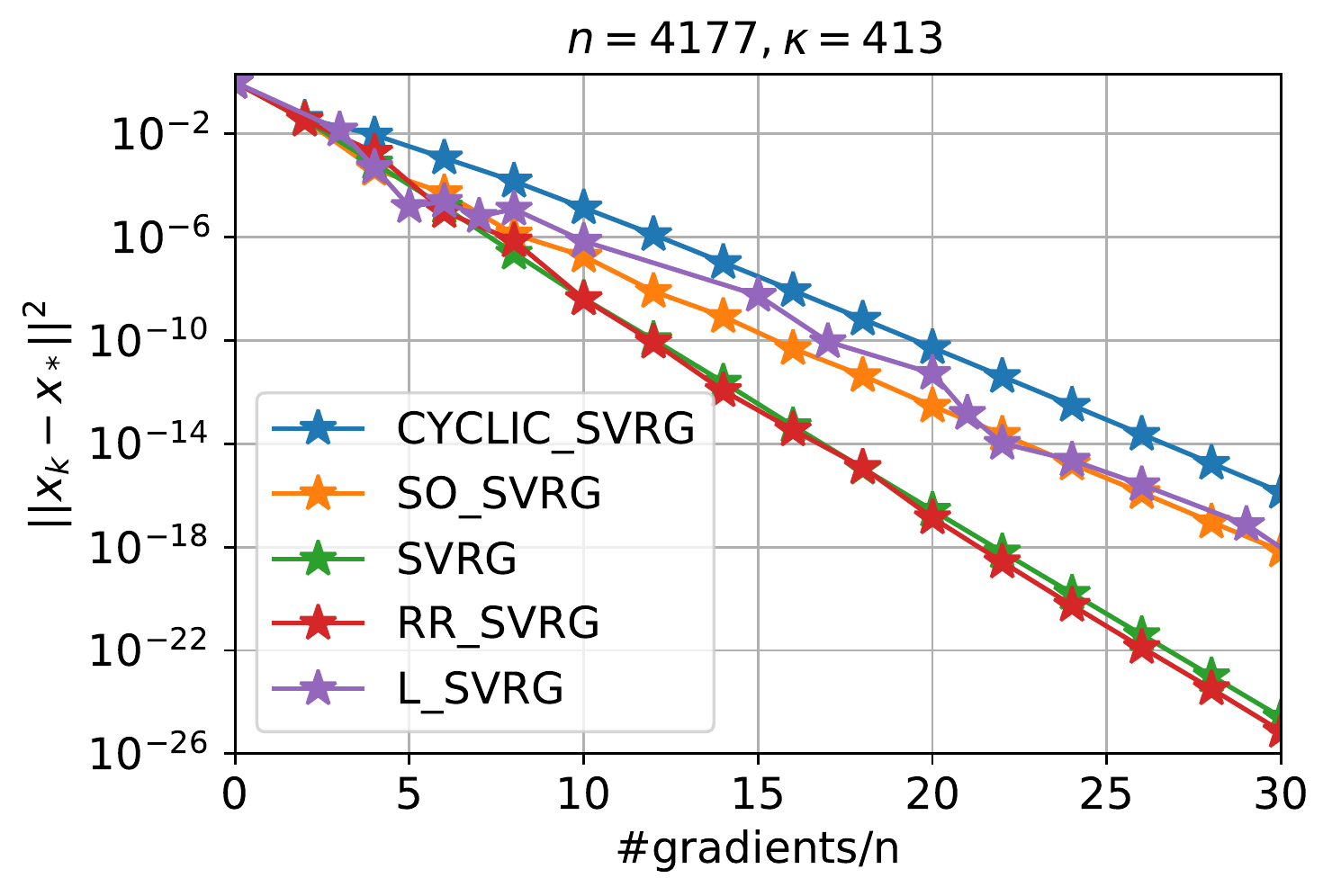}&
			\includegraphics[scale=0.26]{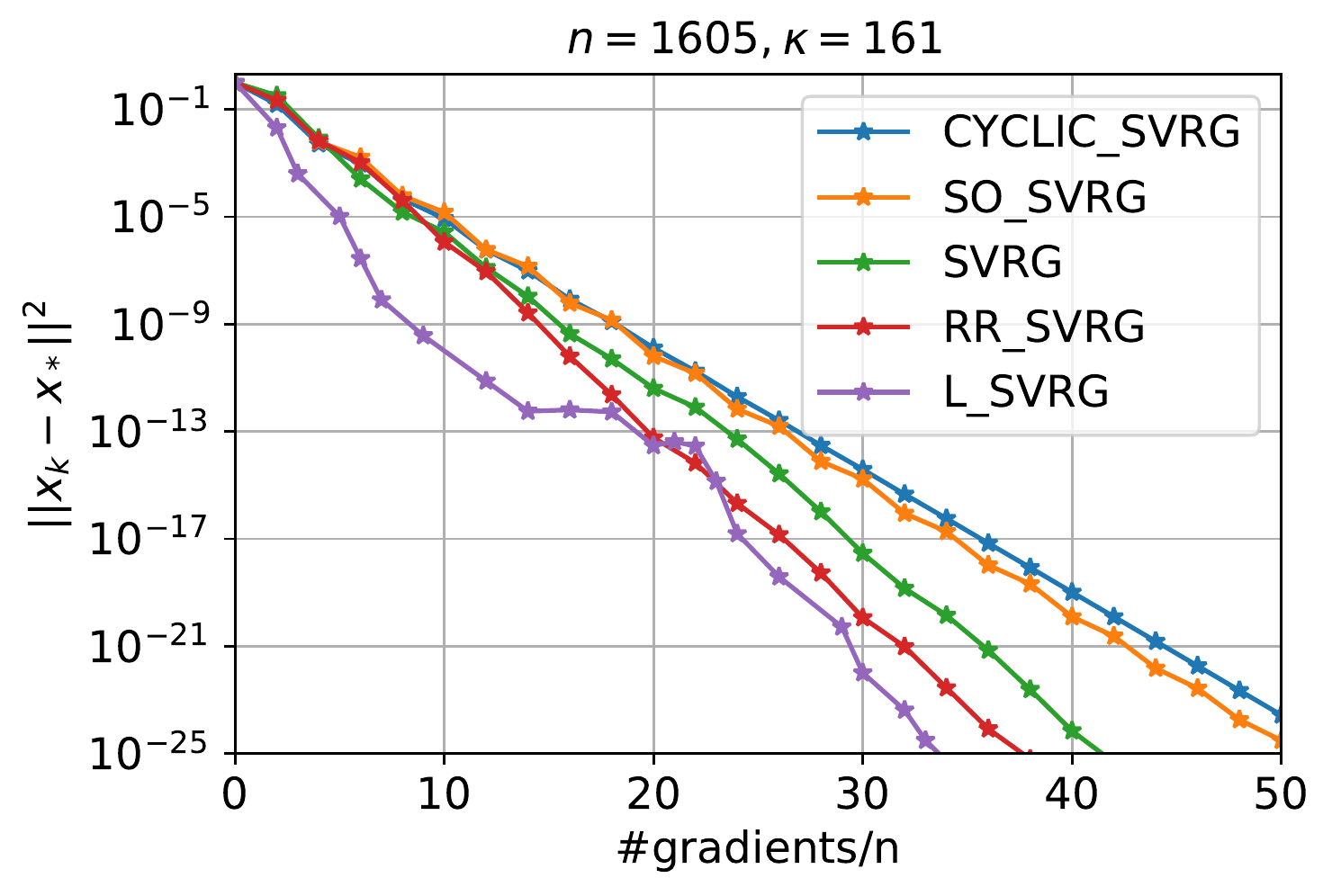}&
			\includegraphics[scale=0.26]{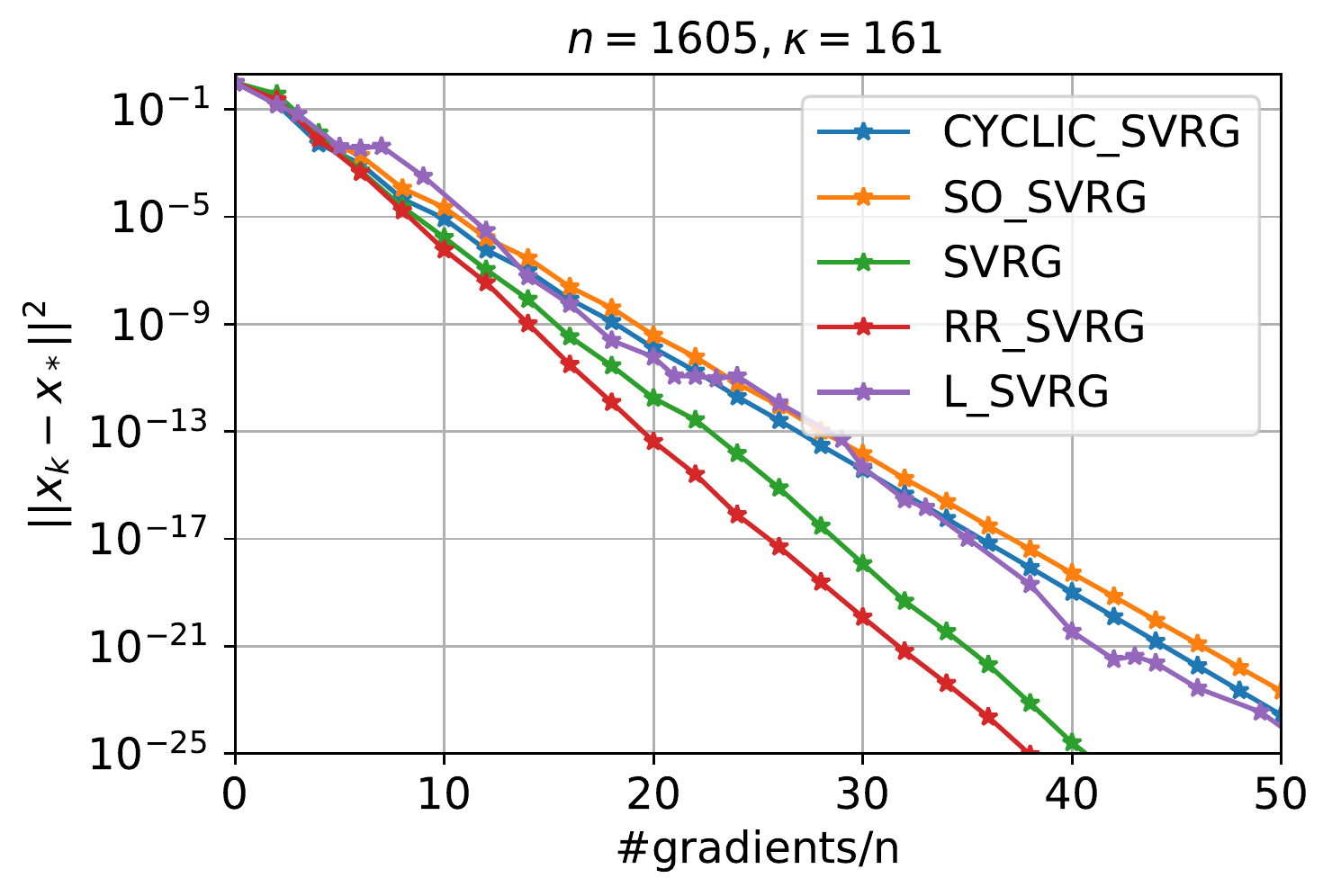}
		\end{tabular}
		\caption{Comparison of SVRG versions: SVRG, L-SVRG, RR-SVRG, CYCLIC-SVRG and SO-SVRG on \texttt{abalone} and \texttt{a1a} datasets. For each dataset we run 5 experiments and take best errors (1st and 3rd plots) and average errors (2nd and 4th plots) for each algorithm.}
		\label{fig:svrg_types}
		\end{figure*}

In our experiments we solve the regularized ridge regression problem, which has the form \eqref{eq:main_finite_sum} with 
\begin{align*}\squeeze	
f_i(x) = \frac{1}{2}\|A_{i,:}x-y_i\|^2 + \frac{\lambda}{2} \|x\|^2,   
\end{align*}
where $A \in \mathbb{R}^{n \times d}, y \in \mathbb{R}^n$ and $\lambda>0$ is a regularizer. Note that this problem is strongly convex and satisfies the Assumptions \ref{L-smooth} for $L = \max_i \|A_{i,:}\|^2 + \lambda $ and $\mu = \frac{\lambda_{\min} (A^\top A)}{n} + \lambda$, where $\lambda_{\min}$ is the smallest eigenvalue. To have a tighter bound on the L-smoothness constant we normalize rows of the data matrix $A$. We use datasets from open LIBSVM corpus~\cite{chang2011libsvm}. In the plots $x$-axis is the number of single data gradient computation divided by $n$, and $y$-axis is the normalized error of the argument $\frac{\|x_k - x_*\|^2}{\|x_0 - x_*\|^2}$. In the appendix you can find details and additional experiments.

\subsection{Random Reshuffling with Variance Reduction}
In this subsection, we compare RR from \cite{mishchenko2020random} and our Algorithm~\ref{alg:RR_VR}. For the Variance Reduced Random Reshuffling algorithm we choose the control update probability $p=0.9$, $p=0.95$ and $p=1$ (RR-SVRG). In our setting, for each algorithm, we choose optimal stepsizes using the grid search. From the plots on Figure~\ref{fig:rr_vs_rrvr}, we can see that the variance reduction indeed works - RR after some iterations starts to oscillate, while RR-VR version converges to the optimum for all choices of $p$. We also can see that depending on the problem, different values of $p$ could be optimal.

\subsection{RR-SVRG vs RR-SAGA}
In this experiment, we compare RR-SVRG and RR-SAGA under an academic setting, i.e. we choose the steps that are suggested by theory. For RR-SVRG we take the stepsize $\gamma = \frac{1}{\sqrt{2} L n}$ when $n \geq \frac{2L}{\mu} \frac{1}{1 - \frac{\mu}{\sqrt{2}L}}$ and $\gamma = \frac{1}{2 \sqrt{2} L n} \sqrt{\frac{\mu}{L}}$ otherwise, and for RR-SAGA $\gamma = \frac{\mu}{11 L^2 n}$. We can see that RR-SVRG outperforms RR-SAGA in terms of the number of epochs and the number of gradient computations. Although the cost of iteration of RR-SVRG is twice higher than RR-SAGA, the larger stepsize significantly impacts the total complexity. In addition, RR-SAGA needs $O(nd)$ extra storage to maintain the table of gradients, which makes RR-SAGA algorithm hard to use in the big data regime.

\subsection{Variance Reduced Random Reshuffling Algorithms}
This section compares the variance reduced algorithms with and without random reshuffling: SAGA, RR-SAGA, SVRG, L-SVRG and RR-SVRG. For each algorithm, we choose its optimal stepsizes using the grid search. To make algorithms reasonable to compare in SVRG, we set the length of the inner loop $m=n$, in L-SVRG the control update probability is $1/n$. Also, we consider only the uniform sampling version of SVRG and L-SVRG. We can see the results on Figure \ref{fig:saga_svrg_rrvr}. We can see that the variance reduced algorithms perform well on this experiment, and there is no obvious leader. However, note that for SAGA and RR-SAGA, we need to have an additional $O(nd)$ space to store the table of the gradients, which is a serious issue in the big data regime.

\subsection{Different versions of SVRG}
In this section, we compare different types of SVRG algorithm: SVRG, L-SVRG, RR-SVRG, SO-SVRG and CYCLIC-SVRG. For each algorithm we run five experiments with different random seeds with optimal stepsizes found by grid search, then we plot the best of the errors (first and third plots) and the average of the errors (second and fourth plots) on Figure~\ref{fig:svrg_types}. We can see that RR-SVRG in average outperforms other algorithms, while in some random cases L-SVRG can perform better. Also, we can see that SO-SVRG is better than CYCLIC-SVRG that coincides with theoretical findings. If the sampling in each epoch is problematic, one can shuffle data once before the training.
\begin{figure*}[h]
	\centering
	\begin{tabular}{ccc}
		\includegraphics[scale=0.35]{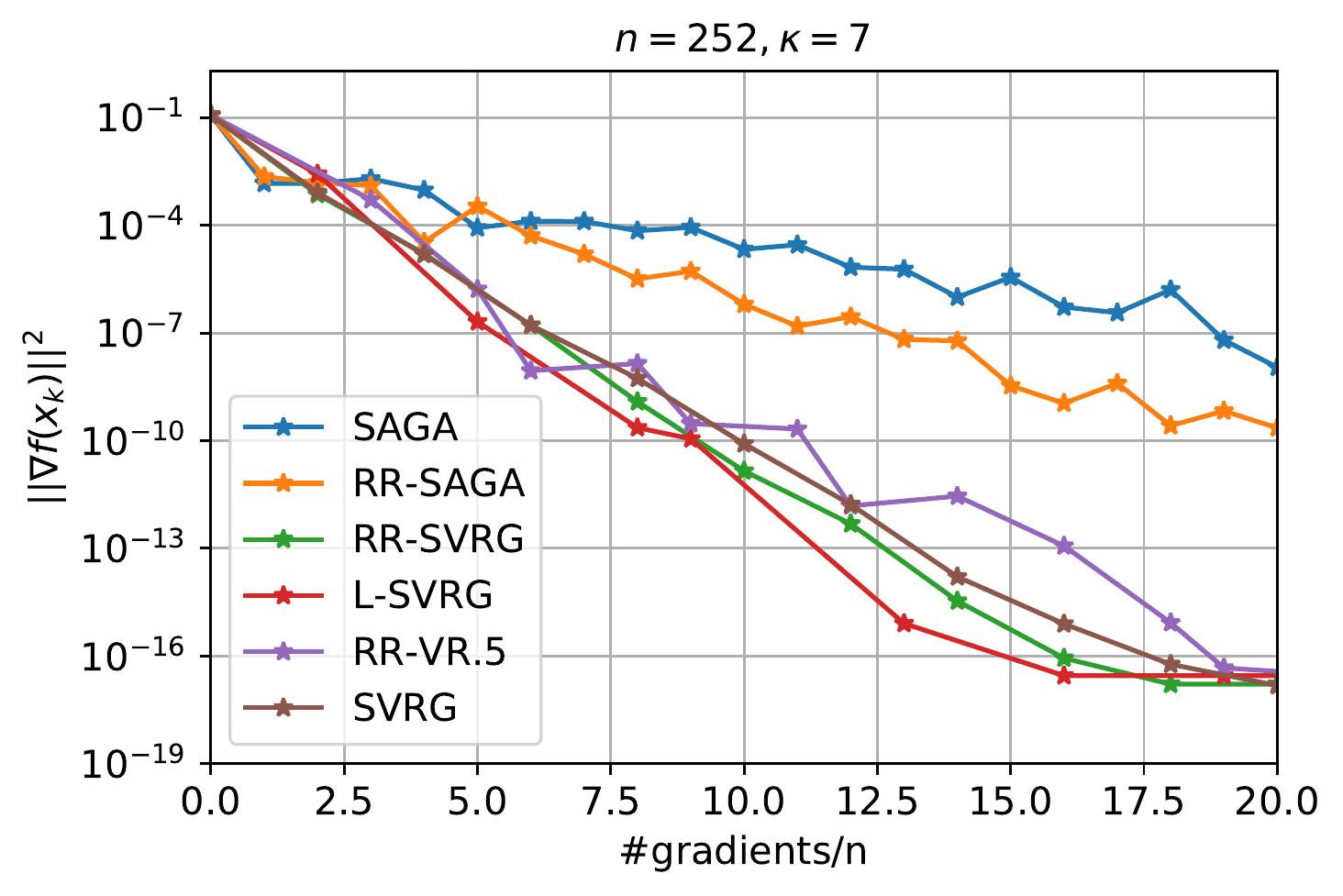} &
		\includegraphics[scale=0.35]{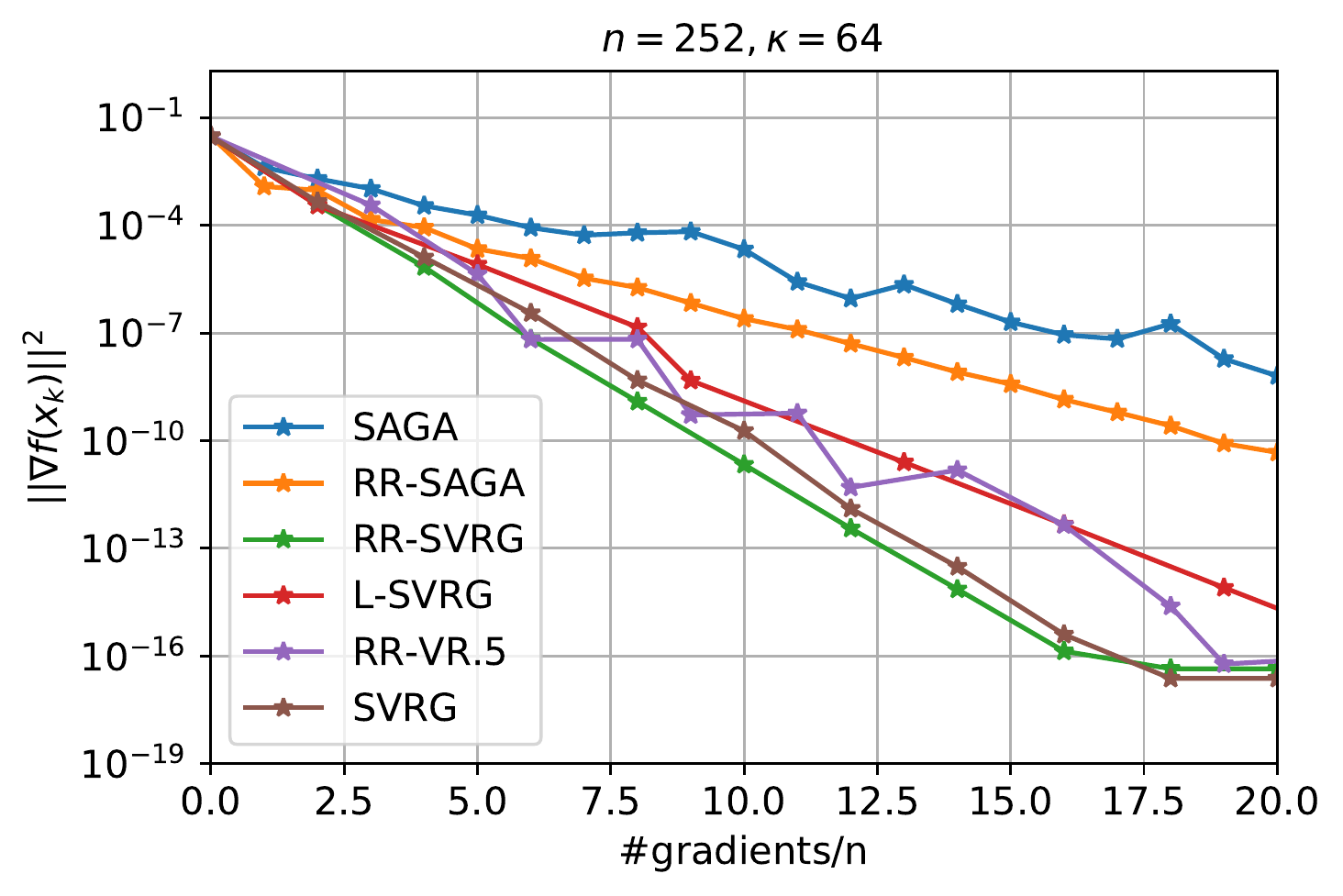} &
		
		\includegraphics[scale=0.35]{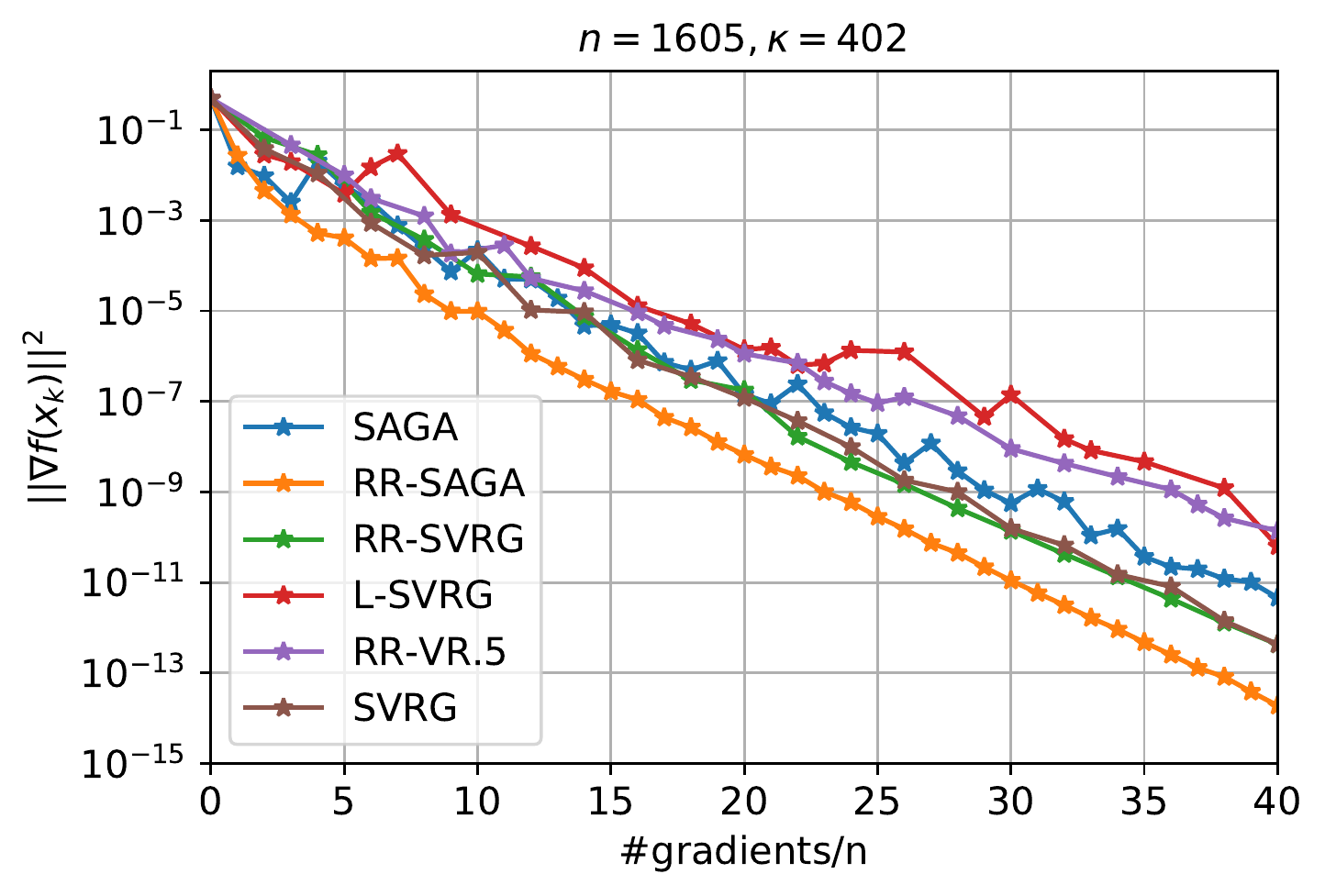} \\
		\includegraphics[scale=0.35]{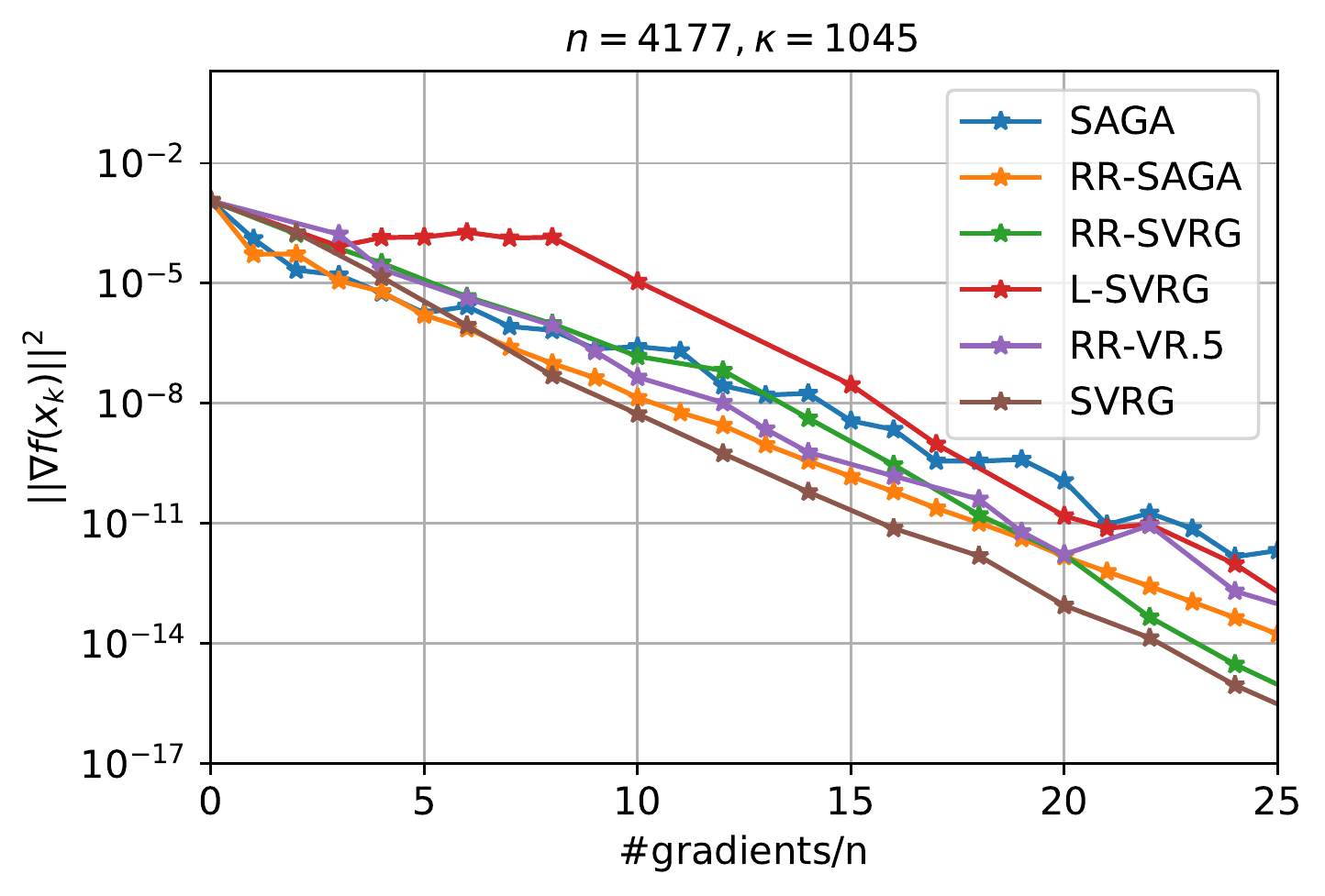} &
		\includegraphics[scale=0.35]{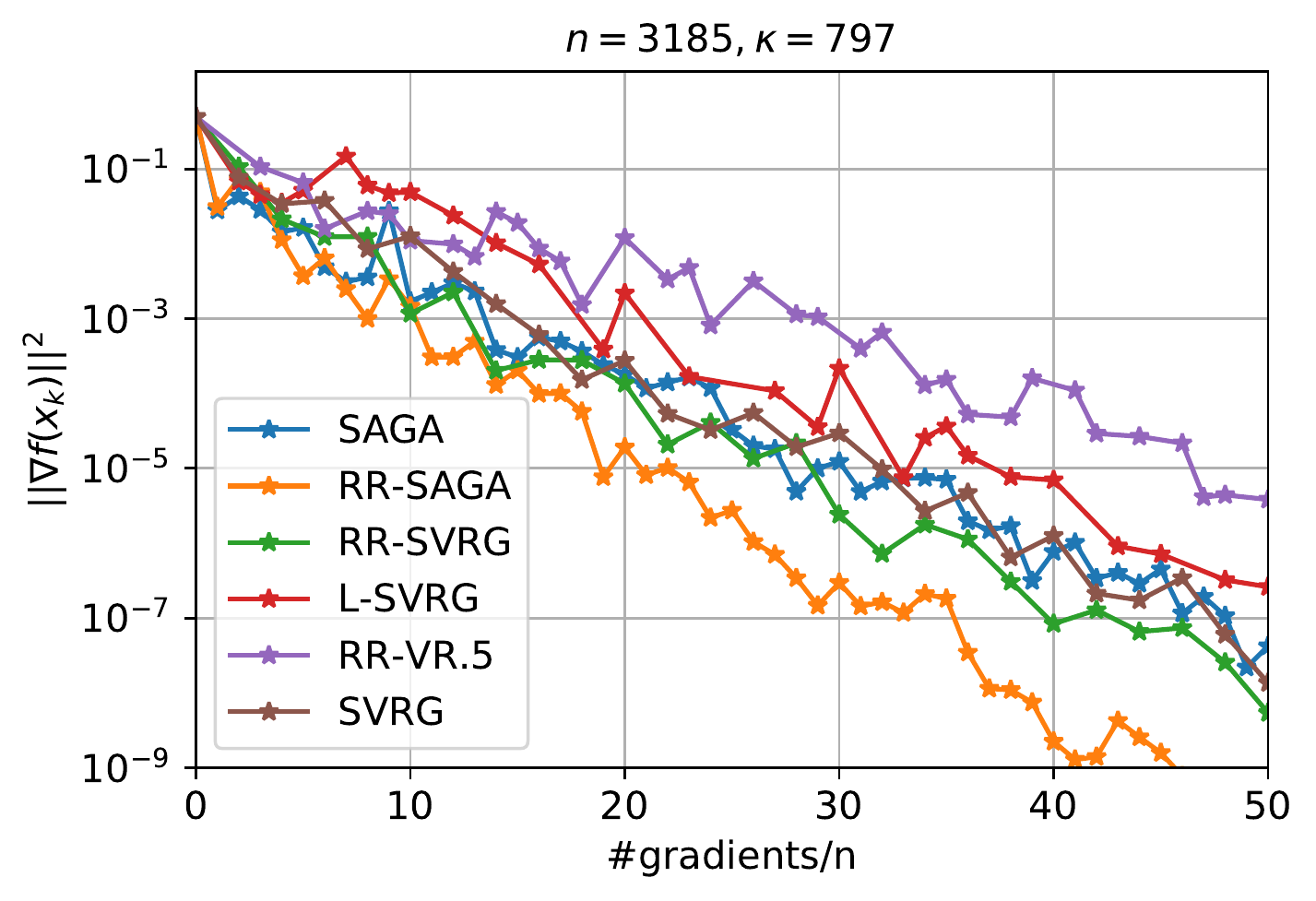} &
		\includegraphics[scale=0.35]{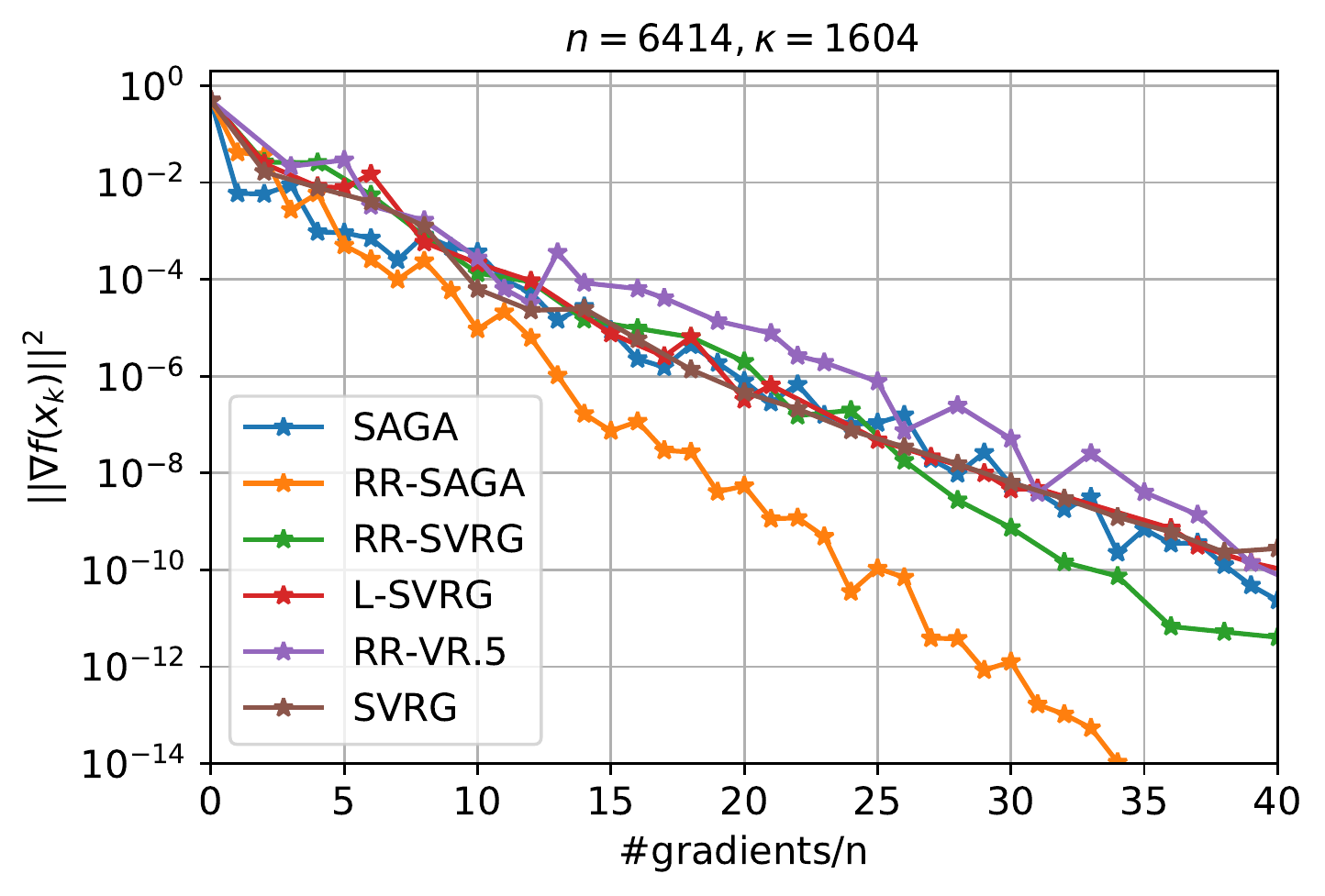}

	\end{tabular}
	\caption{Comparison of SAGA, RR-SAGA, RR-SVRG, L-SVRG, SVRG and RR-VR with $p=0.5$ (RR-VR.5) with optimal stepsizes on \texttt{bodyfat} dataset with different regularization constants (upper left and middle), \texttt{a1a} (upper right), \texttt{abalone} (lower left), \texttt{a3a} (lower middle) and \texttt{a5a} (lower right).}
	\label{fig:logistic_saga_svrg_rrvr}
\end{figure*}

\subsection{Experiments with logistic regression}

We also run experiments for the regularized logistic regression problem; i.e., for problem \eqref{eq:main_finite_sum} with
\begin{align*}
\squeeze
 f(x) = \frac{1}{n} \sum \limits_{i=1}^n  \log \left(1+\exp(-y_i a_i^\top x) \right)+ \frac{\lambda}{2} \|x\|^2.
\end{align*}
Note that the problem is $L$-smooth and $\mu$-strongly convex for $L = \frac{1}{4 n} \lambda_{\max} (A^\top A) + \lambda$, and $\mu = \lambda$. In these experiments (also in the ridge regression experiments) when we choose optimal stepsize, we choose the best one among $\{\frac{1}{L}, \frac{1}{2L}, \frac{1}{3L}, \frac{1}{5L}, \frac{1}{10L}\}$. For the logistic regression we do not have an explicit formula for the optimum $x_*$ as in the ridge regression, thus in this case we compare the norm of the gradients instead. In Figure~\ref{fig:logistic_saga_svrg_rrvr} we can see the performance of the variance reduced algorithms: SAGA, RR-SAGA, SVRG, L-SVRG, RR-VR and RR-SVRG.

\section{Conclusion}
In this paper, we consider variance-reduced algorithms under random reshuffling. Our results are predominantly theoretical because these algorithms are already widely used in practice and show excellent work. We have provided a first-of-its-kind analysis of SVRG in random reshuffling and cyclic mode. We have proposed a new approach for analysis using inner product reformulation, which leads to better rates. Experimental results confirm our theoretical discoveries. Thus, we receive a deeper theoretical understanding of these algorithms' work, and we hope that this will inspire researchers to develop further and analyze these methods. The understanding of variance reduction mechanism is essential to construct accelerated versions for stochastic algorithms. We also believe that our theoretical results can be applied to other aspects of machine learning, leading to improvements in state of the art for current or future applications.

\nocite{langley00}

\bibliography{biblio}
\bibliographystyle{icml2021}
	\clearpage

%\appendix
\onecolumn
\appendix

\part*{Appendix}
\section{Basic Facts}\label{seca1}
\begin{proposition}
	Let $f : \mathbb{R}^d \to \mathbb{R}$ be continuously differentiable and let $L\geq 0$. Then the following statements are equivalent:
	\begin{itemize}
		\item $f$ is $L$-smooth,
		\item $2 D_{f}(x, y) \leq L\|x-y\|^{2} \text { for all } x, y \in \mathbb{R}^{d}$,
		\item $\langle\nabla f(x)-\nabla f(y), x-y\rangle \leq L\|x-y\|^{2} \text { for all } x, y \in \mathbb{R}^{d}$.
	\end{itemize}
\end{proposition}

\begin{proposition}
	Let $f : \mathbb{R}^d \to \mathbb{R}$ be continuously differentiable and let $\mu\geq 0$. Then the following statements are equivalent:
	\begin{itemize}
		\item $f$ is $\mu$-strongly convex,
		\item $2 D_{f}(x, y) \geq \mu\|x-y\|^{2} \text { for all } x, y \in \mathbb{R}^{d}$,
		\item $\langle\nabla f(x)-\nabla f(y), x-y\rangle \geq \mu\|x-y\|^{2} \text { for all } x, y \in \mathbb{R}^{d}$.
	\end{itemize}
\end{proposition}
Note that the $\mu = 0$ case reduces to convexity.

\begin{proposition}
		Let $f : \mathbb{R}^d \to \mathbb{R}$ be continuously differentiable and $L > 0$. Then the following statements are equivalent:
		\begin{itemize}
			\item $f$ is convex and $L$-smooth
			\item $0 \leq 2 D_{f}(x, y) \leq L\|x-y\|^{2} \text { for all } x, y \in \mathbb{R}^{d}$,
			\item $\frac{1}{L}\|\nabla f(x)-\nabla f(y)\|^{2} \leq 2 D_{f}(x, y) \text { for all } x, y \in \mathbb{R}^{d}$,
			\item $\frac{1}{L}\|\nabla f(x)-\nabla f(y)\|^{2} \leq\langle\nabla f(x)-\nabla f(y), x-y\rangle \text { for all } x, y \in \mathbb{R}^{d}$.
		\end{itemize}
\end{proposition}

\begin{proposition}[Jensen’s inequality]
	Let $f: \mathbb{R}^{d} \to \mathbb{R}$ be a convex function, $x_1,\ldots,x_m \in \mathbb{R}^{d}$ and $\lambda_1,\ldots, \lambda_m$ be nonnegative real numbers adding up to 1. Then  
	$$f\left(\sum_{i=1}^{m} \lambda_{i} x_{i}\right) \leq \sum_{i=1}^{m} \lambda_{i} f\left(x_{i}\right).$$
\end{proposition}

\begin{proposition}
	For all $a, b \in \mathbb{R}^{d}$  and $t > 0$ the following inequalities hold
	\begin{align*}
	\langle a, b\rangle &\leq \frac{\|a\|^{2}}{2 t}+\frac{t\|b\|^{2}}{2},\\
	\|a+b\|^{2} &\leq 2\|a\|^{2}+2\|b\|^{2},\\
	\frac{1}{2}\|a\|^{2}-\|b\|^{2} &\leq\|a+b\|^{2}.
	\end{align*}

\end{proposition}

\section{From Rate to Iteration Complexity}
\label{compl_lemma}

We use the following standard result often in the paper. We include the statement and proof, for completeness. Assume that some algorithm satisfies the following recursion for some $q\in (0,1)$ and all $t\geq 0$:
	\begin{align*}
\mathbb{E} \left[ \|x_t - x_* \|^2 \right] \leq \left( 1 - q \right)^t \|x_0 - x_*\|^2.
\end{align*}
Then for any $\varepsilon>0$ we have
\begin{align*}
T \geq \frac{1}{q} \ln \left(\frac{1}{\varepsilon}\right) \qquad \Longrightarrow \qquad \mathbb{E} \left[ \left\|x^{T}-x^{*}\right\|^{2} \right] \leq \varepsilon\left\|x^{0}-x^{*}\right\|^{2}.
\end{align*}

\begin{proof}
	We know that $e^{q} \geq 1+q, \forall q \in \mathbb{R}, \text { thus } e^{-q} \geq 1-q, \forall q \in(0,1).$ Since the function $\ln$ is increasing over $\mathbb{R}_{+}$, it follows that
	\begin{align*}
	-q \geq \ln (1-q), \forall q \in(0,1),
	\end{align*}
	thus for $t\geq 0$, we get
	\begin{align*}
		-t q \geq t \ln \left(1-q\right).
	\end{align*}
	Now if we have $T$ such that 
	$$T\geq \frac{1}{q}\ln\left(\frac{1}{\varepsilon}\right),$$
	which is equivalent to
	$$-T\cdot q \leq \ln (\varepsilon),$$
	we obtain
	$$T \ln \left(1-q\right) \leq \ln (\varepsilon).$$
	Taking exponential on both sides to get
	$$0<\left(1-q\right)^{T} \leq \varepsilon.$$
	Finally, we have 
	$$\mathbb{E} \left[ \left\|x^{T}-x^{*}\right\|^{2} \right] \leq\left(1-q\right)^{T}\left\|x^{0}-x^{*}\right\|^{2} \leq \varepsilon\left\|x^{0}-x^{*}\right\|^{2}.$$
	This leads to
	$$T \geq \frac{1}{q} \ln \left(\frac{1}{\varepsilon}\right) \Longrightarrow\mathbb{E}\left\|x^{T}-x^{*}\right\|^{2} \leq \varepsilon\left\|x^{0}-x^{*}\right\|^{2}.$$
\end{proof}
\section{Proof of Proposition~\ref{prop-reform}}
	Assume that each $f_i$ is $\mu$-strongly convex (convex) and $L$-smooth. Then $\tilde{f}$ defined as 
\begin{equation}
\tilde{f}_i(x) = f_i(x)+\left\langle a_i,x \right\rangle
\end{equation}
is $\mu$-strongly convex (convex) and $L$-smooth.
\begin{proof}
	Let us compute Bregman divergence with respect to the new function $\tilde{f}_i(x):$
	\begin{align*}
	D_{\tilde{f}_i}(x, y) = \tilde{f}_i(x)-\tilde{f}_i(y)-\langle\nabla \tilde{f}_i(y), x-y\rangle.
	\end{align*}
	Note that $\nabla \tilde{f}_i(y) = \nabla f_i(y)+a_i$. Now we have 
	\begin{align*}
	D_{\tilde{f}_i}(x, y) &= \tilde{f}_i(x)-\tilde{f}_i(y)-\langle\nabla \tilde{f}_i(y), x-y\rangle\\
	&= f_i(x)+\left\langle a_i,x \right\rangle - \left( f_i(y)+\left\langle a_i,y \right\rangle\right) - \langle \nabla f_i(y)+a_i, x-y\rangle\\
	&=f_i(x)+\left\langle a_i,x \right\rangle - f_i(y)-\left\langle a_i,y \right\rangle - \langle \nabla f_i(y), x-y\rangle - \langle a_i, x-y \rangle\\
	&=f_i(x)+\left\langle a_i,x \right\rangle - f_i(y)-\left\langle a_i,y \right\rangle - \langle \nabla f_i(y), x-y\rangle - \langle a_i, x \rangle+\langle a_i, y \rangle\\
	&=f_i(x) - f_i(y) - \langle \nabla f_i(y), x-y\rangle\\
	& = D_{f_i}(x, y).
	\end{align*}
	Since the Bregman divergence is not changed, the new function $\tilde{f}_i(x)$ has the same properties ($\mu$-strong convexity or convexity and $L$-smoothness) as the initial function $f_i(x)$. 
		\end{proof}
		
\section{Proof of Lemma~\ref{main_lemma_lemma}}
	Assume that each $f_i$ is $L$-smooth and convex. If we apply the linear perturbation reformulation~\eqref{reform} 
using vectors of the form $ a_i = -\nabla f_{\pi_i} (y_t)+\nabla f (y_t)$, then the variance of reformulated problem satisfies the following inequality:
\begin{equation}
	\tilde{\sigma}_{*}^{2}=\frac{1}{n} \sum \limits_{i=1}^{n}\left\|\nabla \tilde{f}_{i}\left(x_{*}\right)\right\|^{2}\leq 4L^2\|y_t-x_*\|^2.
\end{equation}
\begin{proof}
	Let us start from the definition:
\begin{align}
\notag
\tilde{\sigma}_{*}^2 &= \frac{1}{n}\sum_{i=1}^{n}\| \nabla \tilde{f}_i(x_*) \|^2\\\notag
&=\frac{1}{n}\sum_{i=1}^{n} \|\nabla f_i(x_*)-\nabla f_i(y_t)+\nabla f(y_t) \|^2\\\notag
&=\frac{1}{n}\sum_{i=1}^{n} \|\nabla f_i(x_*)-\nabla f_i(y_t)+\nabla f(y_t) - \nabla f(x_*) \|^2.\\\notag
\end{align}
Using Young's inequality, we get
\begin{align}\notag
\tilde{\sigma}_{*}^2&\leq \frac{1}{n}\sum_{i=1}^{n} \left(2\|\nabla f_i(y_t)-\nabla f_i(x_*)\|^2+2\|\nabla f(y_t) - \nabla f(x_*) \|^2\right)\\\notag
&\leq \frac{1}{n}\sum_{i=1}^{n} 4L_i D_{f_i}(y_t,x_*) +\frac{1}{n}\sum_{i=1}^{n} 4L D_{f}(y_t,x_*)\\\notag
&\leq 4LD_f(y_t,x_*)+4LD_f(y_t,x_*)\\\notag
&=8LD_f(y_t,x_*)\\\notag
&\leq 4 L^2\|y_t-x_*\|^2.
\end{align}
\end{proof}
\section{Analysis of Algorithms~\ref{alg:RRSVRG} and~\ref{alg:SOSVRG}}
\subsection{Proof of Theorems~\ref{th1} and~\ref{th2} }

	Suppose that each $f_i$ is convex, $f$ is $\mu$-strongly convex, and Assumption~\ref{L-smooth}
	holds. Then provided the stepsize satisfies $\gamma \leq \frac{1}{2\sqrt{2} L n}\sqrt{\frac{\mu}{L}},$
	the iterates generated by RR-SVRG (Algorithm~\ref{alg:RRSVRG}) or by SO-SVRG (Algorithm~\ref{alg:SOSVRG}) satisfy
	\begin{align*}
	\mathbb{E} \left[ \|x_T - x_* \|^2 \right] \leq \left( 1 - \frac{\gamma n \mu}{2} \right)^T \|x_0 - x_*\|^2.
	\end{align*}

	Suppose that each $f_i$ is convex, $f$ is $\mu$-strongly convex and Assumption~\ref{L-smooth} holds. Additionally assume we are in the ``big data'' regime characterized by $n \geq \frac{2 L}{\mu} \cdot \frac{1}{1-\frac{\mu}{\sqrt{2} L}}$. Then provided the stepsize satisfies $\gamma \leq \frac{1}{\sqrt{2}Ln},$
	the iterates generated by RR-SVRG (Algorithm~\ref{alg:RRSVRG}) or by SO-SVRG (Algorithm~\ref{alg:SOSVRG}) satisfy
	\begin{align*}
	\mathbb{E} \left[ \|x_T - x_* \|^2 \right] \leq \left( 1 - \frac{\gamma n \mu}{2} \right)^T \|x_0 - x_*\|^2.
	\end{align*}
	\begin{proof}
		 We start from Lemma 3 in paper of~\citet{mishchenko2020random}. 

		 \textbf{Lemma 3.\label{Lemma3}} Assume that functions $f_1, \ldots, f_n$ are convex and that Assumption~\ref{L-smooth} is satisfied. If
		 Random Reshuffling or Shuffle-Once is run with a stepsize satisfying
		 $\gamma\leq\frac{1}{\sqrt{2}Ln}$, then
		 \begin{align*}
		 			 \mathbb{E}\left[\left\|x_{t+1}-x_{*}\right\|^{2}\right] \leq \mathbb{E}\left[\left\|x_{t}-x_{*}\right\|^{2}\right]-2 \gamma n \mathbb{E}\left[f\left(x_{t+1}\right)-f(x_{*})\right]+\frac{\gamma^{3} L n^{2} \sigma_{*}^{2}}{2}.
		 \end{align*}
Now we can apply to the reformulated problem~\eqref{reform}. Using strong convexity we obtain
		 \begin{align*}
\mathbb{E}\left[\left\|x_{t+1}-x_{*}\right\|^{2}\mid x_t\right] &\leq \left\|x_{t}-x_{*}\right\|^{2}-2 \gamma n \mathbb{E}\left[f\left(x_{t+1}\right)-f(x_{*})\mid x_t\right]+\frac{\gamma^{3} L n^{2} \tilde{\sigma}_{*}^{2}}{2}\\
&\leq\left\|x_{t}-x_{*}\right\|^{2}- \gamma n \mu\mathbb{E}\left[\left\|x_{t+1}-x_{*}\right\|^{2}\mid x_t\right] +\frac{\gamma^{3} L n^{2} \tilde{\sigma}_{*}^{2}}{2}.
\end{align*}
For Algorithms~\ref{alg:RRSVRG} and~\ref{alg:SOSVRG} we update $y_t = x_t$ after each epoch, and this leads to
\begin{align*}
\mathbb{E}\left[\left\|x_{t+1}-x_{*}\right\|^{2}\mid x_t\right] &\leq \frac{1}{1+\gamma \mu n}\left(\left\|x_{t}-x_{*}\right\|^{2}+\frac{\gamma^{3} L n^{2} \tilde{\sigma}_{*}^{2}}{2}\right)\\
&\leq \frac{1}{1+\gamma \mu n}\left(\left\|x_{t}-x_{*}\right\|^{2}+\frac{\gamma^{3} L n^{2} \cdot 4 L^2\|y_t-x_*\|^2}{2}\right)\\
&= \frac{1}{1+\gamma \mu n}\left(\left\|x_{t}-x_{*}\right\|^{2}+2\gamma^{3} n^{2} L^3\|x_t-x_*\|^2\right)\\
&= \frac{1}{1+\gamma \mu n}\left(1+2\gamma^{3} n^{2} L^3\right)\|x_t-x_*\|^2.\\
\end{align*}
We can use the tower property to obtain
\begin{equation*}
\mathbb{E}\left[\left\|x_{t+1}-x_{*}\right\|^{2}\right] \leq \frac{1+2 \gamma^{3} L^3 n^{2} }{1+\gamma \mu n} \mathbb{E}\left[\left\|x_{t}-x_{*}\right\|^{2}\right].
\end{equation*}
If this inequality $\frac{1+2 \gamma^{3} L^3 n^{2} }{1+\gamma \mu n} \leq 1 - \frac{\gamma n \mu}{2}$ is correct, we can unroll the recursion and obtain
\begin{equation*}
	\mathbb{E} \left[ \|x_T - x_* \|^2 \right] \leq \left( 1 - \frac{\gamma n \mu}{2} \right)^T \|x_0 - x_*\|^2.
\end{equation*}
Now we need to solve the following inequality:
\begin{align*}
\frac{1+2 \gamma^{3} L^3 n^{2} }{1+\gamma \mu n} \leq 1 - \frac{\gamma n \mu}{2}.
\end{align*}
Let us simplify it:
\begin{align*}
 1+2 \gamma^{3} L^3 n^{2}  &\leq 1 + \frac{\gamma n \mu}{2} - \frac{\gamma^2 n^2 \mu^2}{2} \\
2 \gamma^{2} L^3 n^{2}  &\leq  \frac{n \mu}{2} - \frac{\gamma n^2 \mu^2}{2}   \\
2 \gamma^{2} L^3 n  &\leq  \frac{\mu}{2} - \frac{\gamma n \mu^2}{2}   \\
2 \gamma^{2} L^3 n + \frac{\gamma n \mu^2}{2}  &\leq  \frac{\mu}{2}.   \\
\end{align*}
Now as $\gamma \leq \frac{1}{2\sqrt{2} L n}\sqrt{\frac{\mu}{L}}$, we have
\begin{align*}
	2\cdot\frac{1}{8L^2n^2}\cdot\frac{\mu}{L}L^3n+\frac{1}{2\sqrt{2}Ln}\sqrt{\frac{\mu}{L}}\cdot\frac{n\mu^2}{2}&\leq \frac{\mu}{2}\\
	\frac{1}{4n}\mu + \frac{1}{4\sqrt{2}}\frac{\mu}{L}\sqrt{\frac{\mu}{L}}\mu &\leq \frac{\mu}{2}\\
		\frac{1}{4n} + \frac{1}{4\sqrt{2}}\frac{\mu}{L}\sqrt{\frac{\mu}{L}} &\leq \frac{1}{2}.
\end{align*}
It is true since $n\geq 1$ and $\mu \leq L$. This ends proof of Theorem~\ref{th1}.

Now let us use the biggest step-size allowed by the Lemma 3 in Section~\ref{Lemma3}. Let us utilize $\gamma \leq \frac{1}{\sqrt{2}Ln}:$
\begin{align*}
2\cdot \frac{1}{2L^2n^2}L^3n+\frac{1}{\sqrt{2}Ln} \cdot \frac{n\mu^2}{2}&\leq \frac{\mu}{2}\\
\frac{L}{n}+\frac{\mu}{2}\cdot\frac{\mu}{\sqrt{2}L}&\leq \frac{\mu}{2}.
\end{align*}
This leads to
\begin{align*}
\frac{L}{n}&\leq \frac{\mu}{2} - \frac{\mu}{2}\cdot\frac{\mu}{\sqrt{2}L}=\frac{\mu}{2}\left(1- \frac{\mu}{\sqrt{2}L}\right) \end{align*}
and
\begin{align*}
\frac{1}{n}\leq \frac{\mu}{2L}\left(1- \frac{\mu}{\sqrt{2}L}\right) \quad \Rightarrow \quad 
n\geq \frac{2 L}{\mu} \cdot \frac{1}{1-\frac{\mu}{\sqrt{2} L}}.
\end{align*}
That is the end of Theorem~\ref{th2}.
	\end{proof}
The proofs of Corollary~\ref{corollary1} and~\ref{corollary2} is a simple application of the lemma from Section~\ref{compl_lemma}.
\subsection{Proof of Theorem~\ref{th3}}
	Suppose that the functions $f_1, \ldots, f_n$ are $\mu$-strongly convex and Assumption~\ref{L-smooth} holds. Fix constant $0<\delta<1$. If the stepsize satisfies $\gamma\leq\frac{\delta}{L}\sqrt{\frac{\mu}{2nL}}$ and if number of functions is sufficiently big, $$n>\log\left(\frac{1}{1-\delta^2}\right)\cdot\left(\log\left(\frac{1}{1-\gamma\mu}\right)\right)^{-1},$$ then the iterates generated by RR-SVRG (Algorithm~\ref{alg:RRSVRG}) or by SO-SVRG (Algorithm~\ref{alg:SOSVRG}) satisfy
\begin{align*}
\mathbb{E} \left[ \|x_T - x_* \|^2 \right] \leq \left( \left(1 - \gamma \mu\right)^n +\delta^2 \right)^T \|x_0 - x_*\|^2.
\end{align*}
\begin{proof}
	We start from Theorem 1 in paper of~\citet{mishchenko2020random}:
	$$\mathbb{E}\left[\left\|x_{t+1}-x_{*}\right\|^{2}\mid x_t \right] \leq(1-\gamma \mu)^{n} \left\|x_{t}-x_{*}\right\|^{2}+2 \gamma^{2} \sigma_{\text {Shuffle }}^{2}\left(\sum_{i=0}^{n-1}(1-\gamma \mu)^{i}\right).$$
	Using Preposition 1 in paper of~\citet{mishchenko2020random}:
	$$\frac{\gamma \mu n}{8} \sigma_{*}^{2} \leq \sigma_{\text {Shuffle }}^{2} \leq \frac{\gamma L n}{4} \sigma_{*}^{2},$$
we have 
$$\mathbb{E}\left[\left\|x_{t+1}-x_{*}\right\|^{2}\mid x_t \right] \leq(1-\gamma \mu)^{n} \left\|x_{t}-x_{*}\right\|^{2}+ \frac{\gamma^3 L n}{2} \sigma_{*}^{2}\left(\sum_{i=0}^{n-1}(1-\gamma \mu)^{i}\right). $$
Using property of geometric progression we can have an upper bound for the sum $\sum_{i=0}^{n-1}(1-\gamma \mu)^{i}\leq \frac{1}{\gamma\mu}$:
$$\mathbb{E}\left[\left\|x_{t+1}-x_{*}\right\|^{2}\mid x_t \right] \leq(1-\gamma \mu)^{n} \left\|x_{t}-x_{*}\right\|^{2}+ \frac{\gamma^2 L n}{2\mu} \sigma_{*}^{2}. $$
Now we can apply Lemma~\ref{main_lemma_lemma} and using $y_t = x_t$ we have the following inequality:
\begin{align*}
\mathbb{E}\left[\left\|x_{t+1}-x_{*}\right\|^{2}\mid x_t \right] &\leq(1-\gamma \mu)^{n} \left\|x_{t}-x_{*}\right\|^{2}+ \frac{2\gamma^2 L^3 n}{\mu}\|x_t - x_*\|^2\\
\mathbb{E}\left[\left\|x_{t+1}-x_{*}\right\|^{2}\mid x_t \right] &\leq\left((1-\gamma \mu)^{n} + \frac{2\gamma^2 L^3 n}{\mu}\right)\|x_t - x_*\|^2.\\
\end{align*}
Now we utilize tower property:
\begin{align*}
\mathbb{E}\left[\left\|x_{t+1}-x_{*}\right\|^{2}\right] &\leq\left((1-\gamma \mu)^{n} + \frac{2\gamma^2 L^3 n}{\mu}\right)\mathbb{E}\left[\|x_t - x_*\|^2\right].
\end{align*}
Unrolling this recursion we get:
\begin{align*}
\mathbb{E}\left[\left\|x_{T}-x_{*}\right\|^{2}\right] &\leq\left((1-\gamma \mu)^{n} + \frac{2\gamma^2 L^3 n}{\mu}\right)^T\mathbb{E}\left[\|x_0 - x_*\|^2\right].
\end{align*}
Let us use $\gamma\leq\frac{\delta}{L}\sqrt{\frac{\mu}{2nL}}$:
\begin{align*}
\mathbb{E}\left[\left\|x_{T}-x_{*}\right\|^{2}\right] &\leq\left((1-\gamma \mu)^{n} + \frac{\delta^2}{L^2}\frac{\mu}{2nL}\frac{2 L^3 n}{\mu}\right)^T\mathbb{E}\left[\|x_0 - x_*\|^2\right]\\
\mathbb{E}\left[\left\|x_{T}-x_{*}\right\|^{2}\right] &\leq\left((1-\gamma \mu)^{n} +\delta^2\right)^T\mathbb{E}\left[\|x_0 - x_*\|^2\right].
\end{align*}
To get convergence we need:
$$(1-\gamma \mu)^{n} +\delta^2 <1.$$
This leads to the following inequality:
$$n>\log\left(\frac{1}{1-\delta^2}\right)\cdot\left(\log\left(\frac{1}{1-\gamma\mu}\right)\right)^{-1}$$
This ends the proof.
\subsection{Proof of Corollary~\ref{corollary3}}
	Suppose that assumptions in Theorem~\ref{th3} hold. Additionally assume that $n$ is large enough to satisfy the inequality $\delta^2 \leq (1-\gamma\mu)^{\frac{n}{2}}\left(1-(1-\gamma\mu)^{\frac{n}{2}}\right)$. Then the iteration complexity of Algorithms~\ref{alg:RRSVRG} and~\ref{alg:SOSVRG} is 
\begin{equation*}
T = \mathcal{O}\left(\kappa\sqrt{\frac{\kappa}{n}}\log \left(\frac{1}{\varepsilon}\right)\right).
\end{equation*}
\begin{proof}
	Using the additional assumption 
	$\delta^2 \leq (1-\gamma\mu)^{\frac{n}{2}}\left(1-(1-\gamma\mu)^{\frac{n}{2}}\right),$
	we get
	\begin{align*}
	\delta^2+(1-\gamma\mu)^{n} \leq (1-\gamma\mu)^{\frac{n}{2}}.
	\end{align*} 
	Now we can apply Theorem~\ref{th3} and get
\begin{align*}
\mathbb{E} \left[ \|x_T - x_* \|^2 \right] \leq  \left(1 - \gamma \mu\right)^{\frac{nT}{2}} \|x_0 - x_*\|^2.
\end{align*}
Finally, we apply the lemma from Section~\ref{compl_lemma} with $\gamma\leq\frac{\delta}{L}\sqrt{\frac{\mu}{2nL}}$ and get
\begin{equation*}
T = \mathcal{O}\left(\kappa\sqrt{\frac{\kappa}{n}}\log \left(\frac{1}{\varepsilon}\right)\right).
\end{equation*}
\end{proof}
\subsection{Proof of Theorem~\ref{th4}}
	Suppose the functions $f_1, f_2, \ldots, f_n$ are convex and Assumption~\ref{L-smooth} holds. Then for RR-SVRG (Algorithm~\ref{alg:RRSVRG}) or SO-SVRG (Algorithm~\ref{alg:SOSVRG}) with stepsize $\gamma \leq \frac{1}{\sqrt{2}Ln},$ the average iterate $\hat{x}_{T} \eqdef \frac{1}{T} \sum_{t=1}^{T} x_{t}$ satisfies 
\begin{align*}
\mathbb{E}\left[f\left(\hat{x}_{T}\right)-f\left(x_{*}\right)\right] \leq \frac{3\left\|x_{0}-x_{*}\right\|^{2}}{2 \gamma n T}.
\end{align*}
\begin{proof}
We start with Lemma 3 from~\citet{mishchenko2020random}:
\begin{align*}
\mathbb{E}\left[\left\|x_{t+1}-x_{*}\right\|^{2}\mid x_t\right] \leq \left\|x_{t}-x_{*}\right\|^{2} -2 \gamma n \mathbb{E}\left[f\left(x_{t+1}\right)-f\left(x_{*}\right)\mid x_t\right]+\frac{\gamma^{3} Ln^{2} \sigma_{*}^{2}}{2}\\
2 \gamma n \mathbb{E}\left[f\left(x_{t+1}\right)-f\left(x_{*}\right)\mid x_t\right] \leq \left\|x_{t}-x_{*}\right\|^{2}-\mathbb{E}\left[\left\|x_{t+1}-x_{*}\right\|^{2}\mid x_t\right]+\frac{\gamma^{3} L n^{2} \sigma_{*}^{2}}{2}.
\end{align*}
Using Lemma~\ref{main_lemma_lemma} and considering $y_t=x_t$ we have  
$$\tilde{\sigma}_*^2 \leq 8LD_{\tilde{f}}(x_t,x_*).$$
Applying Proposition~\ref{prop-reform} we get 

$$\tilde{\sigma}_*^2 \leq 8LD_{f}(x_t,x_*) = 8L(f(x_t) - f(x_*)).$$
Next, we utilize the inner product reformulation and get 
\begin{align*}
2 \gamma n \mathbb{E}\left[f\left(x_{t+1}\right)-f\left(x_{*}\right)\mid x_t\right] \leq \left\|x_{t}-x_{*}\right\|^{2}-\mathbb{E}\left[\left\|x_{t+1}-x_{*}\right\|^{2}\mid x_t\right]+\frac{\gamma^{3} L n^{2}}{2} \cdot 8L(f(x_t) - f(x_*)).
\end{align*}
Using tower property we have
\begin{align*}
2 \gamma n \mathbb{E}\left[f\left(x_{t+1}\right)-f\left(x_{*}\right)\right] \leq \mathbb{E}\left[\left\|x_{t}-x_{*}\right\|^{2}\right]-\mathbb{E}\left[\left\|x_{t+1}-x_{*}\right\|^{2}\right]+4\gamma^{3} L^2 n^{2} \mathbb{E}\left[f(x_t) - f(x_*)\right].
\end{align*}
Now we subtract from both sides:
\begin{align*}
2 \gamma n \mathbb{E}\left[f\left(x_{t+1}\right)-f\left(x_{*}\right)\right] - 4\gamma^3L^2n^2\mathbb{E}\left[f\left(x_{t+1}\right)-f\left(x_{*}\right)\right]  &\leq \mathbb{E}\left[\left\|x_{t}-x_{*}\right\|^{2}\right]-\mathbb{E}\left[\left\|x_{t+1}-x_{*}\right\|^{2}\right]\\
&\qquad+4\gamma^{3} L^2 n^{2} \mathbb{E}\left[f(x_t) - f(x_*)\right]\\
& \qquad- 4\gamma^3L^2n^2\mathbb{E}\left[f\left(x_{t+1}\right)-f\left(x_{*}\right)\right]\\
\left(2 \gamma n - 4\gamma^3L^2n^2\right)\mathbb{E}\left[f\left(x_{t+1}\right)-f\left(x_{*}\right)\right] &\leq \mathbb{E}\left[\left\|x_{t}-x_{*}\right\|^{2}\right]-\mathbb{E}\left[\left\|x_{t+1}-x_{*}\right\|^{2}\right]\\
& \qquad+ 4\gamma^{3} L^2 n^{2}\left( \mathbb{E}\left[f(x_t) - f(x_*)\right] - \mathbb{E}\left[f\left(x_{t+1}\right)-f\left(x_{*}\right)\right]\right)\\
2 \gamma n\left(1 - 2\gamma^2L^2n\right)\mathbb{E}\left[f\left(x_{t+1}\right)-f\left(x_{*}\right)\right] &\leq \mathbb{E}\left[\left\|x_{t}-x_{*}\right\|^{2}\right]-\mathbb{E}\left[\left\|x_{t+1}-x_{*}\right\|^{2}\right]\\
& \qquad + 4\gamma^{3} L^2 n^{2}\left( \mathbb{E}\left[f(x_t) - f(x_*)\right] - \mathbb{E}\left[f\left(x_{t+1}\right)-f\left(x_{*}\right)\right]\right).
\end{align*}
Summing these inequalities for $t=0,1,\ldots,T-1$ gives
\begin{align*}
	2 \gamma n\left(1 - 2\gamma^2L^2n\right)\sum_{t=0}^{T-1}\mathbb{E}\left[f\left(x_{t+1}\right)-f\left(x_{*}\right)\right] &\leq \sum_{t=0}^{T-1}\left(\mathbb{E}\left[\left\|x_{t}-x_{*}\right\|^{2}\right]-\mathbb{E}\left[\left\|x_{t+1}-x_{*}\right\|^{2}\right]\right)\\
	&\qquad + 4\gamma^{3} L^2 n^{2}\sum_{t=0}^{T-1}\left( \mathbb{E}\left[f(x_t) - f(x_*)\right] - \mathbb{E}\left[f\left(x_{t+1}\right)-f\left(x_{*}\right)\right]\right)\\
	&= \mathbb{E}\left[\left\|x_{0}-x_{*}\right\|^{2}\right] - \mathbb{E}\left[\left\|x_{T}-x_{*}\right\|^{2}\right]\\
	&\qquad+4\gamma^{3} L^2 n^{2}\mathbb{E}\left[f\left(x_{0}\right)-f\left(x_{*}\right)\right] - 4\gamma^{3} L^2 n^{2}\mathbb{E}\left[f\left(x_{T}\right)-f\left(x_{*}\right)\right]\\
	&\leq \mathbb{E}\left[\left\|x_{0}-x_{*}\right\|^{2}\right] +4\gamma^{3} L^2 n^{2}\mathbb{E}\left[f\left(x_{0}\right)-f\left(x_{*}\right)\right]\\
	&\leq \mathbb{E}\left[\left\|x_{0}-x_{*}\right\|^{2}\right] + 2\gamma^{3} L^3 n^{2}\mathbb{E}\left[\left\|x_{0}-x_{*}\right\|^{2}\right]\\
		&= (1+2\gamma^{3} L^3 n^{2})\mathbb{E}\left[\left\|x_{0}-x_{*}\right\|^{2}\right],
\end{align*}
and dividing both sides by $2 \gamma n\left(1 - 2\gamma^2L^2n\right)T$, we get
\begin{align*}
\frac{1}{T}\sum_{t=0}^{T-1}\mathbb{E}\left[f\left(x_{t+1}\right)-f\left(x_{*}\right)\right] &\leq \frac{1+2\gamma^{3} L^3 n^{2}}{1 - 2\gamma^2L^2n}\frac{\left\|x_{0}-x_{*}\right\|^{2}}{2 \gamma nT}.
\end{align*}
Using the convexity of $f$, the average iterate $\hat{x}_{T} \stackrel{\text { def }}{=} \frac{1}{T} \sum_{t=1}^{T} x_{t}$ satisfies
\begin{align*}
\mathbb{E}\left[f\left(\hat{x}_{T}\right)-f\left(x_{*}\right)\right] \leq \frac{1}{T} \sum_{t=1}^{T} \mathbb{E}\left[f\left(x_{t}\right)-f\left(x_{*}\right)\right]\leq \frac{1+2\gamma^{3} L^3 n^{2}}{1 - 2\gamma^2L^2n}\frac{\left\|x_{0}-x_{*}\right\|^{2}}{2 \gamma nT}.
\end{align*}
Let us show that 
$$\frac{1+2\gamma^{3} L^3 n^{2}}{1 - 2\gamma^2L^2n} \leq 3.$$
Applying $\gamma\leq \frac{1}{\sqrt{2}Ln}$ we have 
$$\frac{1+2\frac{1}{2\sqrt{2}L^3n^3}L^3n^2}{1-2\frac{1}{2L^2n^2}L^2n} = \frac{1+\frac{1}{\sqrt{2}n}}{1-\frac{1}{n}}\leq 3.$$
This leads to $4n>6+\sqrt{2}$ and since $n \in \mathbb{N}:n>1$, this inequality holds. Finally, we have 
$$ \mathbb{E}\left[f\left(\hat{x}_{T}\right)-f\left(x_{*}\right)\right] \leq \frac{3\left\|x_{0}-x_{*}\right\|^{2}}{2 \gamma nT}.$$
\end{proof}
This ends the proof. The Corollary~\ref{corollary4} is a simple application of Theorem~\ref{th4} 

\section{Analysis of Algorithm~\ref{alg:CYCLICSVRG}}
\subsection{Proof of Theorem~\ref{th5}}
	Suppose that each $f_i$ is convex function, $f$ is $\mu$-strongly convex function, and Assumption~\ref{L-smooth}
holds. Then provided the stepsize satisfies $\gamma \leq \frac{1}{4 L n}\sqrt{\frac{\mu}{L}},$
the iterates generated by Cyclic SVRG (Algorithm~\ref{alg:CYCLICSVRG}) satisfy
\begin{align*}
\mathbb{E} \left[ \|x_T - x_* \|^2 \right] \leq \left( 1 - \frac{\gamma n \mu}{2} \right)^T \|x_0 - x_*\|^2.
\end{align*}

We start from Lemma 8 in~\citet{mishchenko2020random}
\begin{align}
	\left\|x_{t+1}-x_{*}\right\|^{2} \leq\left\|x_{t}-x_{*}\right\|^{2}-2 \gamma n\left(f\left(x_{t+1}\right)-f\left(x_{*}\right)\right)+\gamma^{3} L n^{3} \sigma_{*}^{2}.
\end{align}
Now we can apply to the reformulated problem~\eqref{reform}. Using strong convexity we obtain
\begin{align*}
\mathbb{E}\left[\left\|x_{t+1}-x_{*}\right\|^{2}\mid x_t\right] &\leq \left\|x_{t}-x_{*}\right\|^{2}-2 \gamma n \mathbb{E}\left[f\left(x_{t+1}\right)-f(x_{*})\mid x_t\right]+\gamma^{3} L n^{2} \tilde{\sigma}_{*}^{2}\\
&\leq\left\|x_{t}-x_{*}\right\|^{2}- \gamma n \mu\mathbb{E}\left[\left\|x_{t+1}-x_{*}\right\|^{2}\mid x_t\right] +\gamma^{3} L n^{3} \tilde{\sigma}_{*}^{2}.
\end{align*}
For algorithms~\ref{alg:RRSVRG} and~\ref{alg:SOSVRG} we update $y_t = x_t$ after each epoch, this leads to
\begin{align*}
\mathbb{E}\left[\left\|x_{t+1}-x_{*}\right\|^{2}\mid x_t\right] &\leq \frac{1}{1+\gamma \mu n}\left(\left\|x_{t}-x_{*}\right\|^{2}+\gamma^{3} L n^{3} \tilde{\sigma}_{*}^{2}\right)\\
&\leq \frac{1}{1+\gamma \mu n}\left(\left\|x_{t}-x_{*}\right\|^{2}+\gamma^{3} L n^{3} \cdot 4 L^2\|y_t-x_*\|^2\right)\\
&= \frac{1}{1+\gamma \mu n}\left(\left\|x_{t}-x_{*}\right\|^{2}+4\gamma^{3} n^{3} L^3\|x_t-x_*\|^2\right)\\
&= \frac{1}{1+\gamma \mu n}\left(1+4\gamma^{3} n^{3} L^3\right)\|x_t-x_*\|^2.\\
\end{align*}
We can use the tower property to obtain
\begin{equation*}
\mathbb{E}\left[\left\|x_{t+1}-x_{*}\right\|^{2}\right] \leq \frac{1+4 \gamma^{3} L^3 n^{3} }{1+\gamma \mu n} \mathbb{E}\left[\left\|x_{t}-x_{*}\right\|^{2}\right].
\end{equation*}
If this inequality $\frac{1+4 \gamma^{3} L^3 n^{3} }{1+\gamma \mu n} \leq 1 - \frac{\gamma n \mu}{2}$ is correct, we can unroll the recursion and obtain
\begin{equation*}
\mathbb{E} \left[ \|x_T - x_* \|^2 \right] \leq \left( 1 - \frac{\gamma n \mu}{2} \right)^T \|x_0 - x_*\|^2.
\end{equation*}
Now we need to solve the following inequality:
\begin{align*}
\frac{1+4 \gamma^{3} L^3 n^{3} }{1+\gamma \mu n} \leq 1 - \frac{\gamma n \mu}{2}.
\end{align*}
Let us simplify it:
\begin{align*}
1+4 \gamma^{3} L^3 n^{3}  &\leq 1 + \frac{\gamma n \mu}{2} - \frac{\gamma^2 n^2 \mu^2}{2} \\
4 \gamma^{2} L^3 n^{3}  &\leq  \frac{n \mu}{2} - \frac{\gamma n^2 \mu^2}{2}   \\
4 \gamma^{2} L^3 n^2  &\leq  \frac{\mu}{2} - \frac{\gamma n \mu^2}{2}   \\
4 \gamma^{2} L^3 n^2 + \frac{\gamma n \mu^2}{2}  &\leq  \frac{\mu}{2}.   \\
\end{align*}
Now as $\gamma \leq \frac{1}{4 L n}\sqrt{\frac{\mu}{L}}$, we have
\begin{align*}
2\cdot\frac{1}{16L^2n^2}\cdot\frac{\mu}{L}L^3n^2+\frac{1}{4Ln}\sqrt{\frac{\mu}{L}}\cdot\frac{n\mu^2}{2}&\leq \frac{\mu}{2}\\
\frac{1}{4}\mu + \frac{1}{8}\frac{\mu}{L}\sqrt{\frac{\mu}{L}}\mu &\leq \frac{\mu}{2}\\
\frac{1}{4} + \frac{1}{8}\frac{\mu}{L}\sqrt{\frac{\mu}{L}} &\leq \frac{1}{2}.
\end{align*}
It is true since $n\geq 1$ and $\mu \leq L$. This ends proof of Theorem~\ref{th5}.
\end{proof}
The proofs of Corollary~\ref{corollary5} is a simple application of the lemma from Section~\ref{compl_lemma}.
\subsection{Proof of Theorem~\ref{th6}}
	Suppose the functions $f_1, f_2, \ldots, f_n$ are convex and Assumption~\ref{L-smooth} hold.s Then for Algorithm~\ref{alg:CYCLICSVRG} with a stepsize $\gamma \leq \frac{1}{2\sqrt{2}Ln}$, the average iterate $\hat{x}_{T} \eqdef \frac{1}{T} \sum_{j=1}^{T} x_{j}$ satisfies 
\begin{align*}
\mathbb{E}\left[f\left(\hat{x}_{T}\right)-f\left(x_{*}\right)\right] \leq \frac{2\left\|x_{0}-x_{*}\right\|^{2}}{ \gamma n T}.
\end{align*}
We start with Lemma 8 from~\citet{mishchenko2020random}:
\begin{align*}
\mathbb{E}\left[\left\|x_{t+1}-x_{*}\right\|^{2}\mid x_t\right] \leq \left\|x_{t}-x_{*}\right\|^{2} -2 \gamma n \mathbb{E}\left[f\left(x_{t+1}\right)-f\left(x_{*}\right)\mid x_t\right]+\gamma^{3} Ln^{3} \sigma_{*}^{2}\\
2 \gamma n \mathbb{E}\left[f\left(x_{t+1}\right)-f\left(x_{*}\right)\mid x_t\right] \leq \left\|x_{t}-x_{*}\right\|^{2}-\mathbb{E}\left[\left\|x_{t+1}-x_{*}\right\|^{2}\mid x_t\right]+\gamma^{3} L n^{3} \sigma_{*}^{2}.
\end{align*}
Using Lemma~\ref{main_lemma_lemma} and considering $y_t=x_t$, we have  
$$\tilde{\sigma}_*^2 \leq 8LD_{\tilde{f}}(x_t,x_*).$$
Applying Proposition~\ref{prop-reform} we get 
$$\tilde{\sigma}_*^2 \leq 8LD_{f}(x_t,x_*) = 8L(f(x_t) - f(x_*)).$$
Next, we utilize the inner product reformulation and get 
\begin{align*}
2 \gamma n \mathbb{E}\left[f\left(x_{t+1}\right)-f\left(x_{*}\right)\mid x_t\right] \leq \left\|x_{t}-x_{*}\right\|^{2}-\mathbb{E}\left[\left\|x_{t+1}-x_{*}\right\|^{2}\mid x_t\right]+\gamma^{3} L n^{3} \cdot 8L(f(x_t) - f(x_*)).
\end{align*}
Using tower property we have
\begin{align*}
2 \gamma n \mathbb{E}\left[f\left(x_{t+1}\right)-f\left(x_{*}\right)\right] \leq \mathbb{E}\left[\left\|x_{t}-x_{*}\right\|^{2}\right]-\mathbb{E}\left[\left\|x_{t+1}-x_{*}\right\|^{2}\right]+8\gamma^{3} L^2 n^{3} \mathbb{E}\left[(f(x_t) - f(x_*))\right].
\end{align*}
Now we subtract from both sides:
\begin{align*}
2 \gamma n \mathbb{E}\left[f\left(x_{t+1}\right)-f\left(x_{*}\right)\right] - 8\gamma^3L^2n^3\mathbb{E}\left[f\left(x_{t+1}\right)-f\left(x_{*}\right)\right]  &\leq \mathbb{E}\left[\left\|x_{t}-x_{*}\right\|^{2}\right]-\mathbb{E}\left[\left\|x_{t+1}-x_{*}\right\|^{2}\right]\\
&\qquad+8\gamma^{3} L^2 n^{3} \mathbb{E}\left[(f(x_t) - f(x_*))\right]\\
&\qquad - 8\gamma^3L^2n^3\mathbb{E}\left[f\left(x_{t+1}\right)-f\left(x_{*}\right)\right]\\
\left(2 \gamma n - 8\gamma^3L^2n^3\right)\mathbb{E}\left[f\left(x_{t+1}\right)-f\left(x_{*}\right)\right] &\leq \mathbb{E}\left[\left\|x_{t}-x_{*}\right\|^{2}\right]-\mathbb{E}\left[\left\|x_{t+1}-x_{*}\right\|^{2}\right]\\
& \qquad+ 8\gamma^{3} L^2 n^{3}\left( \mathbb{E}\left[f(x_t) - f(x_*)\right] - \mathbb{E}\left[f\left(x_{t+1}\right)-f\left(x_{*}\right)\right]\right)\\
2 \gamma n\left(1 - 4\gamma^2L^2n^2\right)\mathbb{E}\left[f\left(x_{t+1}\right)-f\left(x_{*}\right)\right] &\leq \mathbb{E}\left[\left\|x_{t}-x_{*}\right\|^{2}\right]-\mathbb{E}\left[\left\|x_{t+1}-x_{*}\right\|^{2}\right]\\
& \qquad+ 8\gamma^{3} L^2 n^{3}\left( \mathbb{E}\left[f(x_t) - f(x_*)\right] - \mathbb{E}\left[f\left(x_{t+1}\right)-f\left(x_{*}\right)\right]\right).
\end{align*}
Summing these inequalities for $t=0,1,\ldots,T-1$ gives
\begin{align*}
2 \gamma n\left(1 - 4\gamma^2L^2n^2\right)\sum_{t=0}^{T-1}\mathbb{E}\left[f\left(x_{t+1}\right)-f\left(x_{*}\right)\right] &\leq \sum_{t=0}^{T-1}\left(\mathbb{E}\left[\left\|x_{t}-x_{*}\right\|^{2}\right]-\mathbb{E}\left[\left\|x_{t+1}-x_{*}\right\|^{2}\right]\right)\\
&\qquad + 8\gamma^{3} L^2 n^{3}\sum_{t=0}^{T-1}\left( \mathbb{E}\left[f(x_t) - f(x_*)\right] - \mathbb{E}\left[f\left(x_{t+1}\right)-f\left(x_{*}\right)\right]\right)\\
&= \mathbb{E}\left[\left\|x_{0}-x_{*}\right\|^{2}\right] - \mathbb{E}\left[\left\|x_{T}-x_{*}\right\|^{2}\right]\\
&\qquad+8\gamma^{3} L^2 n^{3}\mathbb{E}\left[f\left(x_{0}\right)-f\left(x_{*}\right)\right] - 8\gamma^{3} L^2 n^{3}\mathbb{E}\left[f\left(x_{T}\right)-f\left(x_{*}\right)\right]\\
&\leq \mathbb{E}\left[\left\|x_{0}-x_{*}\right\|^{2}\right] +8\gamma^{3} L^2 n^{3}\mathbb{E}\left[f\left(x_{0}\right)-f\left(x_{*}\right)\right]\\
&\leq \mathbb{E}\left[\left\|x_{0}-x_{*}\right\|^{2}\right] + 4\gamma^{3} L^3 n^{3}\mathbb{E}\left[\left\|x_{0}-x_{*}\right\|^{2}\right]\\
&= (1+4\gamma^{3} L^3 n^{3})\mathbb{E}\left[\left\|x_{0}-x_{*}\right\|^{2}\right],
\end{align*}
and dividing both sides by $2 \gamma n\left(1 - 4\gamma^2L^2n^2\right)T$, we get
\begin{align*}
\frac{1}{T}\sum_{t=0}^{T-1}\mathbb{E}\left[f\left(x_{t+1}\right)-f\left(x_{*}\right)\right] &\leq \frac{1+4\gamma^{3} L^3 n^{3}}{1 - 4\gamma^2L^2n^2}\frac{\left\|x_{0}-x_{*}\right\|^{2}}{2 \gamma nT}.
\end{align*}
Using the convexity of $f$, the average iterate $\hat{x}_{T} \stackrel{\text { def }}{=} \frac{1}{T} \sum_{t=1}^{T} x_{t}$ satisfies
\begin{align*}
\mathbb{E}\left[f\left(\hat{x}_{T}\right)-f\left(x_{*}\right)\right] \leq \frac{1}{T} \sum_{t=1}^{T} \mathbb{E}\left[f\left(x_{t}\right)-f\left(x_{*}\right)\right]\leq \frac{1+4\gamma^{3} L^3 n^{3}}{1 - 4\gamma^2L^2n^2}\frac{\left\|x_{0}-x_{*}\right\|^{2}}{2 \gamma nT}.
\end{align*}
Let us show that 
$$\frac{1+4\gamma^{3} L^3 n^{3}}{1 - 4\gamma^2L^2n^2} \leq 4.$$
Applying $\gamma\leq \frac{1}{2\sqrt{2}Ln}$ we have 
$$\frac{1+4\frac{1}{16\sqrt{2}L^3n^3}L^3n^3}{1-4\frac{1}{8L^2n^2}L^2n^2} = \frac{1+\frac{1}{4\sqrt{2}}}{1-\frac{1}{2}}\leq 4.$$
Finally, we have 
$$ \mathbb{E}\left[f\left(\hat{x}_{T}\right)-f\left(x_{*}\right)\right] \leq \frac{2\left\|x_{0}-x_{*}\right\|^{2}}{\gamma nT}.$$
This ends the proof. The Corollary~\ref{corollary6} is a simple application of Theorem~\ref{th6}.
\section{Analysis of Algorithm~\ref{alg:RR_VR}}
\subsection{Proof of Theorem~\ref{th7}}
	Suppose that each $f_i$ is convex, $f$ is $\mu$-strongly convex, and Assumption~\ref{L-smooth}
holds. Then provided the parameters satisfy $n>\kappa$, $\frac{\kappa}{n}<p<1$ and $\gamma \leq \frac{1}{2\sqrt{2}Ln}$, 
the final iterate generated by RR-VR (Algorithm~\ref{alg:RR_VR}) satisfies
\begin{align*}
V_{T} \leq \max \left( q_1,q_2 \right)^{T} V_{0},
\end{align*}
where
\begin{align*}
q_1 = 1-\frac{\gamma \mu n}{4}\left(1-\frac{p}{2}\right), \quad
q_2 = 1-p+\frac{8}{\mu} \gamma^{2} L^{3} n,
\end{align*}
and the Lyapunov function is defined via
\begin{align*}
V_t \eqdef \mathbb{E}\left[\left\|x_{t}-x_{*}\right\|^{2}\right]+\frac{4}{\gamma\mu n}\mathbb{E}\left[\left\|y_{t}-x_{*}\right\|^{2}\right]. 
\end{align*}
\begin{proof}
For the problem $\frac{1}{n}\sum_{i=1}^{n}\tilde{f}_i(x)$ we will use an inequality from \citet{mishchenko2020random}:
\begin{align*}
\mathbb{E}\left[\left\|x_{t+1}-x_{*}\right\|^{2}\mid x_t\right] & \leq \frac{1}{1+\gamma \mu n}\left(\left\|x_{t}-x_{*}\right\|^{2}+\frac{\gamma^{3} L n^{2} \sigma_{*}^{2}}{2}\right) \\
&=\frac{1}{1+\gamma \mu n} \left\|x_{t}-x_{*}\right\|^{2}+\frac{1}{1+\gamma \mu n} \frac{\gamma^{3} L n^{2} \sigma_{*}^{2}}{2} \\
& \leq\left(1-\frac{\gamma \mu n}{2}\right) \left\|x_{t}-x_{*}\right\|^{2}+\frac{\gamma^{3} L n^{2} \sigma_{*}^{2}}{2}.
\end{align*}
Now we apply inequality 
\begin{align*}
\mathbb{E}\left[\left\|x_{t+1}-x_{*}\right\|^{2}\mid x_t, y_t\right] &\leq\left(1-\frac{\gamma \mu n}{2}\right) \left\|x_{t}-x_{*}\right\|^{2}+\frac{\gamma^{3} L n^{2} \sigma_{*}^{2}}{2}\\
&\leq\left(1-\frac{\gamma \mu n}{2}\right) \left\|x_{t}-x_{*}\right\|^{2} + 2\gamma^3L^3n^2\|y_t-x_*\|^2.
\end{align*}
Using tower property we have 
\begin{align*}
\mathbb{E}\left[\left\|x_{t+1}-x_{*}\right\|^{2}\right] &= \mathbb{E}\left[\mathbb{E}\left[\left\|x_{t+1}-x_{*}\right\|^{2}\mid x_t, y_t\right]\right] \\
&\leq\left(1-\frac{\gamma \mu n}{2}\right) \mathbb{E}\left[\left\|x_{t}-x_{*}\right\|^{2}\right] +2\gamma^3L^3n^2\mathbb{E}\left[\|y_t-x_*\|^2\right].
\end{align*}
Now we look at 
\begin{align*}
y_{t+1}=\left\{\begin{array}{ll} y_t & \text{with probability } 1-p \\
x_t & \text{with probability } p
\end{array}\right.
\end{align*}
We get
\begin{align*}
\mathbb{E}\left[\|y_{t+1} - x_*\|^2\mid x_t,y_t\right] = (1-p)\|y_t - x_*\|^2+p\| x_t - x_* \|^2.
\end{align*}
Using tower property 
\begin{align*}
\mathbb{E}\left[\|y_{t+1} - x_*\|^2\right] &= \mathbb{E}\left[\mathbb{E}\left[\|y_{t+1} - x_*\|^2\mid x_t,y_t\right]\right]\\
&= (1-p)\mathbb{E}\left[\|y_t - x_*\|^2\right]+p\mathbb{E}\left[\| x_t - x_* \|^2\right].
\end{align*}
Finally, we have 
\begin{align*}
\mathbb{E}\left[\left\|x_{t+1}-x_{*}\right\|^{2}\right]+M\mathbb{E}\left[\|y_{t+1} - x_*\|^2\right]  & \leq\left(1-\frac{\gamma \mu n}{2}\right) \left\|x_{t}-x_{*}\right\|^{2} +2\gamma^3L^3n^2\mathbb{E}\left[\|y_t-x_*\|^2\right]\\\notag
&\qquad+(1-p)M\mathbb{E}\|y_t - x_*\|^2 +pM\mathbb{E}\| x_t - x_* \|^2.
\end{align*}
Denote $V_{t} = \mathbb{E}\left[\left\|x_{t}-x_{*}\right\|^{2}\right]+M\mathbb{E}\left[\|y_{t} - x_*\|^2\right].$ Using this we obtain
\begin{align*}
V_{t+1}&\leq\left(1-\frac{\gamma \mu n}{2}\right) \mathbb{E}\left[\left\|x_{t}-x_{*}\right\|^{2}\right] +2\gamma^3L^3n^2\mathbb{E}\left[\|y_t-x_*\|^2\right]\\\notag
&\qquad+(1-p)M\mathbb{E}\left[\|y_t - x_*\|^2\right]+pM\mathbb{E}\left[\| x_t - x_* \|^2\right].
\end{align*}
Thus,
\begin{align*}
V_{t+1}&\leq\left(1-\frac{\gamma \mu n}{2}+pM\right) \mathbb{E}\left[\left\|x_{t}-x_{*}\right\|^{2}\right] +\left(1-p+\frac{1}{M}2\gamma^3L^3n^2\right)M\mathbb{E}\left[\|y_t-x_*\|^2\right].
\end{align*}
To have contraction we use 
\begin{align*}
M = \frac{\gamma\mu n}{4}, \qquad \gamma = \frac{1}{2\sqrt{2}Ln}.
\end{align*}
We have the final rate
\begin{align*}
V_{t+1} &\leq \max \left(1-\frac{\gamma \mu n}{4}\left(1-\frac{p}{2}\right),  1-p+\frac{8}{\mu} \gamma^{2} L^{3} n \right)V_t\\
V_{T} &\leq \max\left(1-\frac{\gamma \mu n}{4}\left(1-\frac{p}{2}\right),  1-p+\frac{8}{\mu} \gamma^{2} L^{3} n \right)^TV_0.
\end{align*}
\end{proof} 
The proof of Corollary~\ref{corollary7} is an application of the lemma from Section~\ref{compl_lemma}.
\subsection{Proof of Theorem~\ref{th_last}}
	Suppose that the functions $f_1, \ldots, f_n$ are $\mu$-strongly convex, and that Assumption~\ref{L-smooth} holds. Then for RR-VR (Algorithm~\ref{alg:RR_VR}) with parameters that satisfy $\gamma \leq \frac{1}{2L}\sqrt{\frac{\mu}{2nL}}$, $\frac{1}{2}<\delta<\frac{1}{\sqrt{2}}$, $0<p<1$, and for a sufficiently large number of functions, $n>\log\left(\frac{1}{1-\delta^2}\right)\cdot\left(\log\left(\frac{1}{1-\gamma\mu}\right)\right)^{-1}$, the iterates generated by the RR-VR algorithm satisfy
\begin{align*}	
V_{T} &
\leq \max \left(q_1,q_2\right)^{T} V_{0},	
\end{align*}
where $$  q_1 
= (1-\gamma \mu)^{n}+\delta^2, \quad
q_2 	
= 1-p\left(1-\frac{2\gamma^2L^3n}{\mu\delta^2}\right),$$ and 
$$
V_t \eqdef \mathbb{E}\left[\left\|x_{t}-x_{*}\right\|^{2}\right]+\frac{\delta^2}{p}\mathbb{E}\left[\left\|y_{t}-x_{*}\right\|^{2}\right]. 
$$
\begin{proof}
For the problem $\frac{1}{n}\sum_{i=1}^{n}\tilde{f}_i(x)$ we will use two inequalities from \citet{mishchenko2020random}:
\begin{align*}
\mathbb{E}\left[\left\|x_{t+1}-x_{*}\right\|^{2}\mid x_t\right]  &\leq\left(1-\gamma \mu\right)^n \left\|x_{t}-x_{*}\right\|^{2}+2\gamma^2\sigma_{\text {Shuffle }}^2\left(\sum_{i=0}^{n-1}(1-\gamma\mu)^i\right)\\
\sigma_{\text {Shuffle }}^{2} &\leq \frac{\gamma L n}{4} \sigma_{*}^{2}.
\end{align*}
Using this result, we have 
\begin{align*}
\mathbb{E}\left[\left\|x_{t+1}-x_{*}\right\|^{2}\mid x_t,y_t\right]  &\leq\left(1-\gamma \mu\right)^n \left\|x_{t}-x_{*}\right\|^{2}+\frac{1}{2}\gamma^3Ln\sigma_{*}^2\left(\sum_{i=0}^{n-1}(1-\gamma\mu)^i\right)\\
&\leq\left(1-\gamma \mu\right)^n\left\|x_{t}-x_{*}\right\|^{2}+\frac{1}{\mu}2\gamma^2L^2nL\|y_t - x_*\|^2.
\end{align*}
Using tower property
\begin{align*}
\mathbb{E}\left[\left\|x_{t+1}-x_{*}\right\|^{2}\right] &= \mathbb{E}\left[\mathbb{E}\left[\left\|x_{t+1}-x_{*}\right\|^{2}\mid x_t,y_t\right]\right]\\
&  \leq\left(1-\gamma \mu\right)^n\mathbb{E}\left[\left\|x_{t}-x_{*}\right\|^{2}\right]+\frac{1}{\mu}2\gamma^2LnL^2\mathbb{E}\left[\|y_t - x_*\|^2\right].
\end{align*}
Now we look at 
\begin{align*}
y_{t+1}=\left\{\begin{array}{ll} y_t & \text{with probability } 1-p \\
x_t & \text{with probability } p
\end{array}\right..
\end{align*}
Thus,
\begin{align*}
\mathbb{E}\left[\|y_{t+1} - x_*\|^2\mid x_t,y_t\right] = (1-p)\|y_t - x_*\|^2+p\| x_t - x_* \|^2.
\end{align*}
Using tower property 
\begin{align*}
\mathbb{E}\left[\|y_{t+1} - x_*\|^2\right] &= \mathbb{E}\left[\mathbb{E}\left[\|y_{t+1} - x_*\|^2\mid x_t,y_t\right]\right]\\
&= (1-p)\mathbb{E}\left[\|y_t - x_*\|^2\right]+p\mathbb{E}\left[\| x_t - x_* \|^2\right].
\end{align*}
Denote $V_{t} = \mathbb{E}\left[\left\|x_{t}-x_{*}\right\|^{2}\right]+M\mathbb{E}\left[\|y_{t} - x_*\|^2\right]$ and we have
\begin{align*}
V_{t+1} &= \mathbb{E}\left[\left\|x_{t+1}-x_{*}\right\|^{2}\right]+M\mathbb{E}\left[\|y_{t+1} - x_*\|^2\right]\\
& \leq\left(1-\gamma \mu\right)^n\mathbb{E}\left[\left\|x_{t}-x_{*}\right\|^{2}\right]+\frac{2}{\mu}\gamma^2L^3n\mathbb{E}\left[\|y_t - x_*\|^2\right]+(1-p)M\mathbb{E}\left[\|y_t - x_*\|^2\right]+pM\mathbb{E}\left[\| x_t - x_* \|^2\right]\\
&\leq \left(\left(1-\gamma \mu\right)^n+pM\right)\mathbb{E}\left[\left\|x_{t}-x_{*}\right\|^{2}\right] +\left((1-p)+\frac{2\gamma^2L^3n}{\mu M}\right)M\mathbb{E}\left[\| x_t - x_* \|^2\right]\\
&\leq\max\left(\left(\left(1-\gamma \mu\right)^n+pM\right),\left((1-p)+\frac{2\gamma^2L^3n}{\mu M}\right)\right)V_t.
\end{align*}
Unrolling the recusrion we have 
\begin{align*}
V_T \leq \max\left(\left((1-\gamma\mu)^n+pM\right), \left(1-p + \frac{2\gamma^2L^3n}{\mu M}\right)\right)^TV_0.
\end{align*}
Applying $M = \frac{\delta^2}{p}$ and $\gamma \leq \frac{1}{2L}\sqrt{\frac{\mu}{2nL}}$ we get 
\begin{align*}	
V_{T} &	
\leq \max \left((1-\gamma \mu)^{n}+\delta^2,1-p\left(1-\frac{2\gamma^2L^3n}{\mu\delta^2}\right)\right)^{T} V_{0}.
\end{align*}
\end{proof}
The proof of Corollary~\ref{corollary8} repeats the proof of Corollary~\ref{corollary3}.

\end{document}